%% file: emnlp2023.tex
\pdfoutput=1

\documentclass[11pt]{article}
\usepackage{EMNLP2023}

\usepackage{times}
\usepackage{latexsym}
\usepackage[T1]{fontenc}
\usepackage[utf8]{inputenc}
\usepackage{microtype}
\usepackage{inconsolata}
\usepackage{booktabs}
\usepackage{multirow}
\usepackage{tabularx}
\usepackage{xcolor}
\usepackage{array}
\usepackage{xspace}
\usepackage{graphicx}
\usepackage{pgfplots}
\pgfplotsset{compat=1.9,width=3cm}
\usepackage{subcaption}
\usepackage{enumitem}
\usepackage{amsfonts}
\usepackage{amsmath}
\usepackage{hyperref}

\newcommand{\palmii}[1]{\mbox{PaLM 2-#1}}
\newcommand{\fpalmii}[1]{\mbox{F-PaLM 2-#1}}

\def\dataset/{\textit{riSum}}
\def\rouge/{\textsc{Rouge}}
\def\bleurt/{\textsc{Bleurt}}
\def\bartscore/{\textsc{BARTScore}}

\newenvironment{boxed2}
    {\begin{center}
    \begin{tabular}{|p{0.95\textwidth}|}
    \hline\\
    }
    { 
    \\\\\hline
    \end{tabular} 
    \end{center}
    }
\newenvironment{boxed2*}
    {\begin{center}
    \begin{tabular}{|p{0.475\textwidth}|}
    \hline\\
    }
    { 
    \\\\\hline
    \end{tabular} 
    \end{center}
    }

\title{Towards Better Evaluation of Instruction-Following:\\A Case-Study in Summarization}
\author{Ondrej Skopek\thanks{\;\;Correspondence to: \href{mailto:oskopek@google.com}{oskopek@google.com}}\\
Google Research
\And
Rahul Aralikatte\\
Google Research
\And
Sian Gooding\\
Google Research
\And
Victor C\u{a}rbune\\
Google Research}

\begin{document}
\maketitle
\begin{abstract}%
Despite recent advances, evaluating how well large language models (LLMs) follow user instructions remains an open problem. While evaluation methods of language models have seen a rise in prompt-based approaches, limited work on the correctness of these methods has been conducted.
In this work, we perform a meta-evaluation of a variety of metrics to quantify how accurately they measure the instruction-following abilities of LLMs.
Our investigation is performed on grounded query-based summarization by collecting a new short-form, real-world dataset riSum,
containing $300$ document-instruction pairs with $3$ answers each.
All $900$ answers are rated by $3$ human annotators. Using riSum, we analyze the agreement between evaluation methods and human judgment.
Finally, we propose new LLM-based reference-free evaluation methods that
improve upon established baselines and
perform on par with costly reference-based metrics that require high-quality summaries.
\end{abstract}

\begin{figure*}[tb]
\centering
\small
\begin{boxed2}
\textbf{Document}: \\[0.2em]
Lee, You should be receiving a package shortly containing the following: (...)\\
3. {\color{magenta}Assignment and assumption agreement to move the equipment from TurboPark  to the CAED I. There will be one for CAED II as well.} This document is  being reviewed by the bank, so I'm not convinced it is in final form. You  will note that there is an acknowledgement section for GE. {\color{violet}I cut and pasted  from the consent to assignment from the TurboPark documents, but shortened  the whole thing considerably.} Here's that document: \\    
4. {\color{orange}Signature pages (signed by Enron) from the ESA deal, both the facility  agreement and the override letter. Obviously, we need your signature. I will forward the final CA facility agreements to you once again, along with the blacklines against what you initialled.}\\[0.5em]
\textbf{Instructions}:\\[0.2em]
$\;\bullet\;$ Briefly {\color{magenta}describe the purpose of the assignment and assumption agreement} mentioned in the paragraph.\vspace{0.25em}\\
$\;\bullet\;$ Explain the {\color{violet}changes made to the GE acknowledgement section in the context of the TurboPark documents}.\vspace{0.25em}\\
$\;\bullet\;$ Summarize the {\color{orange}final steps regarding the CA facility agreements and signature pages}.\\[0.3em]
\textbf{Answers (for instruction \#1)}:\\[0.2em]
\begin{tabular}{@{}p{4.5em}p{40em}@{}}
\setlength\tabcolsep{0pt}
\fpalmii{S} & The assignment and assumption agreement is to {\color{magenta}move the equipment from TurboPark to the CAED I}. \\
\fpalmii{Sc} & The purpose of the assignment and assumption agreement is to {\color{magenta}move the equipment from TurboPark to the CAED I}. \\
GPT-3.5 &  (...) The purpose of the assignment and assumption agreement {\color{magenta}is not specified}.
\end{tabular}
\vspace{-0.5em}
\end{boxed2}
\caption{Randomly sampled example from \dataset/ (data source: AESLC). Highlighted how GPT-4 transforms parts of the input document into grounded instructions.}
\label{fig:example}
\end{figure*}

\section{Introduction}

Large Language Models (LLMs) have shown human-level performance in many NLP tasks. Recent advances in instruction tuning \citep{ouyang2022training,brown2020language} and alignment
\citep{10.5555/3495724.3495977,zhou2023lima} have dramatically increased the ability of these models to follow instructions. In addition to being used to tackle unseen tasks in zero-shot setups \citep{chung2022scaling}, these models are now also used as surrogates to human annotators, especially for NLG tasks \citep{chiang2023can,wu2023large}, where human evaluations are time-consuming and expensive.

Consider the instruction ``Briefly describe the purpose of the assignment and assumption agreement mentioned in the paragraph'' from Figure~\ref{fig:example}. There are several dimensions to evaluate a generated output on: (i) \textit{Coherence}: whether it is understandable and free of grammatical mistakes, (ii) \textit{Faithfulness}: whether facts in the output are supported by the document, (iii) \textit{Style}: whether specific formatting requirements (lists, brevity, ...) are met, and (iv) \textit{Alignment}: whether it semantically fulfills the instruction.

Analyzing these different facets for each model output increases the cognitive load of annotators,
thereby increasing the likelihood of errors or low-quality evaluations \cite{goyal-etal-2022-falte}.
It also increases the turnaround time and hence annotations become expensive.
An increasingly popular alternative is to ask LLMs to evaluate the generated outputs.
Recent work like \citet{liu2023geval} and \citet{fu2023gptscore} show that LLMs can produce human-like evaluations of text by using clever prompting techniques \cite{wei2023chainofthought,yao2023tree}.
But preliminary studies have shown that LLMs can be inconsistent in their evaluations and can easily be influenced \citep{wang2023large,shen2023large}. 
\citet{gehrmann2023repairing} have also looked at evaluation flaws and have recommended that metric developers should focus on metrics with smaller, but better defined scopes (like instruction-following).

Hence, there is an urgent need for a standard framework to analyze the specifics of instruction-following abilities of LLMs. SummEval \cite{fabbri2021summeval} proposes something similar for vanilla summarization. Doing this for instruction-following can be tricky because we would like to not only evaluate the LLMs as task solvers \textit{``Summarize this document in 20 words or less''}, but also as task evaluators \textit{``Does the summary satisfy the conditions of the instruction?''}. The meta-evaluation framework should be robust and ideally reference-free \citep{liu2023geval}. Reference-free evaluation for text generation has been widely studied \citep{liu-etal-2022-reference,hessel2022clipscore,ke-etal-2022-ctrleval}, but to the best of our knowledge, there has been no prior work on reference-free evaluations for instruction-following.

In this work, we take the first steps towards building such a framework. To make this problem tractable, we choose to limit our scope to the task of query-based summarization. We consider this to be an appropriate initial task since
(i) numerous domains to source documents from exist, (ii) the space of appropriate instructions is broad, while still (iii) maintaining groundedness of both instructions and answers into facts present in the documents. We leave the expansion of the dataset in size and domain/instruction scope to future work.

\paragraph{Contributions}
For this purpose, we release a rated, instructed summarization dataset \dataset/%
\footnote{\,The dataset will be made available at \href{https://goo.gle/risum}{goo.gle/risum}.},
consisting of 900 instruction-summary pairs with 3 human ratings each (Figure~\ref{fig:example}).

We introduce several reference-free evaluation methods which perform on-par with expensive reference-based methods and outperform existing reference-free baselines in terms of correlation with human judgement.

Lastly, we leverage \dataset/ to perform an extensive meta-evaluation, quantifying how well different evaluation methods are able to replace human judgments by statistically ranking model outputs.

\paragraph{Model naming}
In this work, we rely on different LLMs for a variety of tasks. Specifically, we use GPT-3.5 \citep{ouyang2022training}
and GPT-4\footnote{\,OpenAI model id: gpt-4-0314} \citep{openai2023gpt4} models from the GPT LLM family,
and \palmii{S} and \palmii{L} models from the PaLM family \citep{palm2}.
The models are also finetuned on the Flan corpus as described in \citet[Appendix A.2]{palm2}, denoted as \fpalmii{S} and \fpalmii{L}.
Finally, these models are further finetuned using standard methods and data known to improve instruction-following \cite{alpaca}, denoted as \fpalmii{Sc} and \fpalmii{Lc}.

\section{Data Collection}\label{sec:collection}

\subsection{Dataset collection}

\begin{table}[tb]
    \centering
    \small
    \setlength{\tabcolsep}{0.25em}
    \begin{tabular}{@{}p{17em} ccc@{}}
\toprule
Data source & Min & Med & Max \\
\midrule 
AESLC emails \cite{zhang-tetreault-2019-email} & 118 & 172.0 & 469 \\
arXiv abstracts \cite{clement2019arxiv} & 122 & 145.5 & 224 \\
BBC news \cite{xsum-emnlp} & 173 & 272.5 & 473 \\
CNN/DM news \cite{DBLP:conf/nips/HermannKGEKSB15} & 244 & 465.5 & 532 \\
Common Crawl \cite{2019t5} & 127 & 282.5 & 506 \\
ForumSum threads \cite{khalman-etal-2021-forumsum-multi} & 158 & 320.0 & 519 \\
Reddit posts \cite{volske-etal-2017-tl} & 156 & 299.0 & 552 \\
SAMSum dialogues \cite{gliwa-etal-2019-samsum} & 127 & 189.5 & 384 \\  
Task-Oriented dialogues \cite{lee2022sgd} & 161 & 329.5 & 605 \\
Yelp reviews \cite{zhangCharacterlevelConvolutionalNetworks2015} & 119 & 140.5 & 357 \\
\bottomrule
    \end{tabular}
    \caption{Data sources from which \dataset/ is sampled and the minimum (Min), median (Med), and maximum (Max) sampled document length (in words). 10 documents were sampled without replacement from each of the 10 data sources.}
    \label{tab:data-stats}
\end{table}

\paragraph{Data sourcing}%
To create \dataset/, a total of 100 documents are chosen from 10 existing datasets of different domains to ensure the data is as diverse as possible.
The documents are uniformly sampled from each dataset, restricting to documents with a word count between 100 and 500 words (Table~\ref{tab:data-stats}).

\paragraph{Instruction generation}%
To procure instructions for each document, we first evaluate the quality of generations from four models: \fpalmii{Sc}, \fpalmii{Lc}, GPT-3.5, and GPT-4.
We randomly sample 10 documents from the dataset and let each model generate 3 instructions per document.
Each of the 40 ($10\times4$ models) document-instructions pairs was rated ``good'', ``neutral'', or ``bad'' by three evaluators in a side-by-side setting. In this evaluation, GPT-4 outperformed the other models on 6/10 documents, therefore we used it to sample instructions for all documents in the dataset. This results in a total of 300 document-instruction pairs.

\paragraph{Answer generation}%
Subsequently, three different models\footnote{\,\fpalmii{S}, \fpalmii{Sc}, and GPT-3.5. We do not use GPT-4 as it was used to generate the instructions.} are used to generate answers for each of the document-instruction pairs, yielding the final dataset with 900 data points.

\paragraph{Human evaluation}%
Finally, each document-instruction-output triplet individually is evaluated by at least three human annotators. They are asked two questions:
\begin{enumerate}[left=0pt,itemsep=-0.25em,topsep=0.25em,partopsep=0pt]
    \item Does the output follow the instruction? (Y/N).
    \item Rate the output on a scale of 1 to 5. 1 indicates the output does not follow the instruction at all, 5 indicates the instruction is followed strictly.
\end{enumerate}
See Appendix~\ref{app:annotators} for a description of the annotator UI,
Appendix~\ref{app:guidelines} for annotator guidelines, and
Appendix~\ref{app:prompts} for the instruction-generation prompt.

\subsection{Analysis of Human Ratings}

\begin{figure}[tb]
    \hspace{-8px}\resizebox{0.51\textwidth}{!}{\input{rater_local_alpha_histogram.pgf}}
    \caption{Histogram of local Krippendorff $\alpha$ for document-instruction pairs.}
    \label{fig:alpha-hist}
\end{figure}
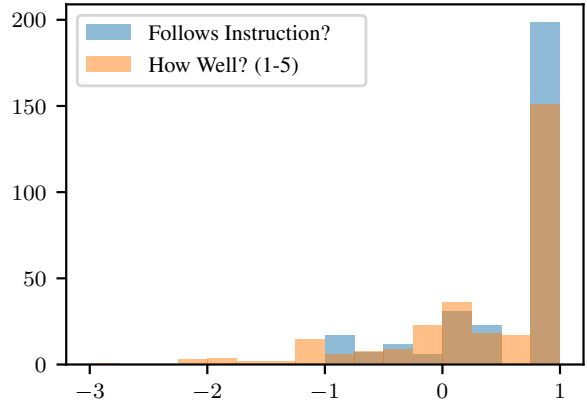

\begin{table}[tb]
\small
\centering
\setlength{\tabcolsep}{0.25em}
\begin{tabular}{@{}l lc p{0.4em} lc@{}}
\toprule
\multirow{2}{*}{$\alpha$ method}& \multicolumn{2}{c}{Follows Instruction?} && \multicolumn{2}{c}{How Well?}\\
\cmidrule{2-3}\cmidrule{5-6}
& Mean$\pm$SE & $\geq50\%$ && Mean$\pm$SE & $\geq50\%$ \\ \midrule
Global$_{n=900}$ & ${54.3}$ & && ${11.4}$ &\\
Local$_{n=295}$  & ${62.1}\pm{3.6}$ & ${67.5}$ && ${31.9}\pm{4.5}$ & ${56.9}$\\
\bottomrule
\end{tabular}
\caption{Krippendorff $\alpha$ values (in $\%$) of \dataset/ human ratings. $\geq50\%$ denotes the $\%$ of pairs where $\alpha \geq 0.5$.}
\label{tab:alphas}
\end{table}

\begin{table*}[tb]
\small
\centering
\begin{subtable}[t]{0.48\textwidth}
\small
\centering
\setlength{\tabcolsep}{0.5em}
\begin{tabular}{@{}cl ccc@{}}
\toprule
Label & Agg. & \fpalmii{S} & \fpalmii{Sc} & GPT-3.5\\
\midrule
\multirow{3}{*}{FI} & Mean & $80.8\pm{1.8}$ & $85.0\pm{1.6}$ & $94.0\pm{1.0}$\\
& Maj. & $81.3\pm{2.2}$ & $86.8\pm{2.0}$ & $96.7\pm{1.0}$\\
& None & $80.9\pm{1.3}$ & $85.0\pm{1.2}$ & $94.0\pm{0.8}$\\
\midrule
\multirow{3}{*}{HW} & Mean & $61.1\pm{1.8}$ & $65.0\pm{1.6}$ & $73.5\pm{1.2}$\\
& Maj. & $59.8\pm{2.2}$ & $64.4\pm{2.1}$ & $74.8\pm{1.5}$\\
& None & $61.1\pm{1.2}$ & $65.1\pm{1.2}$ & $73.6\pm{0.9}$\\
\bottomrule
\end{tabular}
\caption{Average annotator response per answer ($\%$, higher is better).}
\label{tab:model_comparison_answer}
\end{subtable}
\hfill
\begin{subtable}[t]{0.49\textwidth}
\small
\centering
\setlength{\tabcolsep}{0.5em}
\begin{tabular}{@{}cl ccc@{}}
\toprule
Label & Agg. & \fpalmii{S} & \fpalmii{Sc} & GPT-3.5\\
\midrule
\multirow{3}{*}{FI} & Mean & $2.12\pm{0.05}$ & $2.05\pm{0.05}$ & $1.83\pm{0.05}$\\
& Maj. & $2.10\pm{0.05}$ & $2.02\pm{0.05}$ & $1.87\pm{0.05}$\\
& None & $2.12\pm{0.05}$ & $2.05\pm{0.05}$ & $1.83\pm{0.05}$\\
\midrule
\multirow{3}{*}{HW}& Mean & $2.16\pm{0.05}$ &$2.09\pm{0.05}$ &$1.75\pm{0.05}$\\
& Maj. &$2.16\pm{0.05}$ &$2.04\pm{0.05}$ &$1.80\pm{0.05}$\\
& None &$2.16\pm{0.05}$ & $2.09\pm{0.05}$ & $1.75\pm{0.05}$\\
\bottomrule
\end{tabular}
\caption{Model ranking per (doc., instr.) pair (1--3, lower is better).}
\label{tab:model_comparison_docinstr}
\end{subtable}
\caption{Aggregate model quality according to human ratings.
``Mean'' aggregation takes the mean of human ratings for each model output ($n=300$), ``Maj.'' takes the majority vote with ties broken randomly ($n=300$), and ``None'' performs no aggregation ($n=900$).
Averaged across 100,000 runs.
FI is the binary rating ``Follows Instruction?'',
HW is the qualitative rating of ``How Well?''. Ratings are normalized to $0$--$1$ and reported as \%.}
\label{tab:model_comparisons}
\end{table*}

For analyzing annotator agreement (Table~\ref{tab:alphas}), we leverage locally and globally computed Krippendorff $\alpha$ \citep{krippendorff}. For the first boolean question, we use the nominal distance function (indicator function) and for the second ordinal question, we use the interval distance method (squared difference). For local application, we compute a localized $\alpha$ for each document-instruction pair and then aggregate the results over all pairs. We omit 5 document-instruction pairs from the analysis for which the Krippendorff $\alpha$ is not defined because there is no annotator overlap among the 3 ratings for each of the 3 model outputs.

We note that around 67\% of the dataset has high levels of agreement on the first question and 57\% on the second question. The tail of disagreement is long (Figure~\ref{fig:alpha-hist}), but we hypothesize that given the difficulty of rating outputs in these diverse and highly specific texts, disagreements would be non-negligible even with higher replication rates. At the expense of gathering only relative information, ranking two responses against each other instead of rating single responses may help. Given the diversity of domains and instructions, hiring domain experts for future ratings could help increase quality and agreement, whilst also increasing costs.

Additionally, factoring out independent rating dimensions (e.g. language level, factuality) may help quantify common mistakes types in LLM instruction following and identity misalignment areas with respect to human expectations at the expense of a slower and more expensive rating process.

In Table~\ref{tab:model_comparison_answer}, we present aggregate numbers for annotator preferences among the three model outputs.
We explore the mean of ratings, majority consensus votes (ties broken randomly), and a global mean over individual ratings (no aggregation).
In Table~\ref{tab:model_comparison_docinstr}, the three model outputs are ranked for each document-instruction pair and the ranking indices are then averaged across the dataset, with ties broken randomly.
Both tables are averaged over 100,000 runs to eliminate noise from tie-breaking.

\section{Evaluation Methods}

We propose and evaluate several methods that model annotator preferences, focusing our analysis on reference-based vs.~reference-free methods and their effectiveness in different data regimes.

\subsection{Reference-based methods}\label{sec:methods-ref}

Reference-based methods require access to at least one reference answer which can be considered the ``gold standard'' for each document-instruction pair.
Given numerous prior work noting that summaries written by crowd workers exhibit limitations associated with lack of annotator expertise in the domain \citep{gillickliu2010},
especially at narrower tasks like query-based summarization \citep{jiang-etal-2018-effective}, we use LLM-generated references for benchmarking reference-based methods instead.

The requirement of having access to high-quality references fundamentally limits the utility of the methods.
In all our reference-based experiments, we use GPT-4 and \fpalmii{Lc} generated summaries as references.
Since we use GPT-3.5 and \fpalmii{S}, and \fpalmii{Sc} to generate candidate answers for evaluations,
we use larger variants of these models to generate the ``gold'' references,
which ensures that they are generally of higher quality
(see e.g.~Table 19 of \citealp{palm2}).

\paragraph{\bleurt/ (model-based)}
\citet{sellam2020bleurt} take a (candidate, reference) answer pair as input and aim to model semantic similarity between the two texts. In all results below, we use the \bleurt/\textsubscript{20} model \cite{pu2021learning}. In scenarios with multiple reference answers, we take the maximum \bleurt/\textsubscript{20} score across all reference answers.

\paragraph{\rouge/ (n-gram-based)}
\citet{lin2004rouge} also take (candidate, reference) pairs as input and measure n-gram overlap to provide a numerical estimate of how well the candidate resembles the reference. We report the geometric mean of \rouge/\textsubscript{1}, \rouge/\textsubscript{2}, and \rouge/\textsubscript{Lsum} and refer to this method as \rouge/\textsubscript{avg}. Similar to \bleurt/, in a scenario with multiple reference answers, we report the maximum \rouge/\textsubscript{avg} score for a given candidate.

\subsection{Reference-free baseline methods}\label{sec:methods-free}

We investigate popular heuristics (e.g.~length of the generated response) and several LM-based approaches, varying the amount of data used.
Fine-tuning a model on a subset of the collected data would also yield a viable evaluation method, but we leave that for future exploration.

\paragraph{Length-based heuristics}
The simplest reference-free method we use is based on length heuristics. The length of the model output is a common source of bias in human ratings when evaluating the quality of summaries, where longer answers are often preferred over shorter ones, since the former usually contains more information. Therefore, it is a natural baseline for assessing the degree to which the collected ratings suffer from this type of bias. We simply count the words and sentences using NLTK \cite{nltk} and meta-evaluate how they would behave if they were used as a proxy for generated answer quality.

\paragraph{Model-based methods}
We benchmark the following state-of-the-art model-based methods on the \dataset/ dataset: (i) \bartscore/ and \bartscore/\textsubscript{CNN} \citep{bartscore}, and (ii) T5\textsubscript{ANLI} \citep{true}. Both are encoder-decoder Transformer models and have around 400M and 11B parameters respectively.

\subsection{LLM-based reference-free methods}\label{sec:methods-llm}

The following methods depend on an underlying LLM for evaluation. Though we use PaLM 2 models in our experiments, these methods are model agnostic, and any LLM can be used in their place.
For the following methods, we leverage either the base \palmii{S/L} models, or the instruction-tuned \fpalmii{Sc/Lc}.

\paragraph{Constrained Softmax}
We feed the underlying model two prompts: one for the ``Follows instruction? (Y/N)'' question, and another for the ``How well? (1-5)'' question. The prompts used correspond to the task descriptions provided to annotators (Prompts presented in Figure~\ref{fig:scoring-prompt} of Appendix~\ref{app:prompts}).

Instead of sampling tokens to obtain the ratings, we use the model to compute the negative log-likelihood of all the possible rating values (``Yes''/``No'' for the first question,  $\{1,2,3,4,5\}$ for the second question) and pick the most likely token as the rating.
This approach has multiple advantages over generating tokens directly: 
\begin{enumerate}[left=0pt,itemsep=-0.25em,topsep=0.25em,partopsep=0pt]
    \item \textit{Correctness}: The model can never output a rating that is not from the list of options.
    \item \textit{Efficiency}: All our rating values are a single token in the model's vocabulary, which makes the scoring extremely efficient.
    Additionally, repeated sampling is not necessary to obtain a more precise estimate of the model's rating.
    \item \textit{Uncertainty}: By re-normalizing the likelihoods across all rating values, we obtain a rating distribution, which lets us precisely quantify the confidence the model assigns to ratings.
    For an unbiased estimate with respect to the logits, we fix the softmax temperature to 1.
\end{enumerate}
Finally, we return the expected value for each of the question's distributions:
{%
\setlength{\abovedisplayskip}{5pt}
\setlength{\belowdisplayskip}{5pt}
\setlength{\abovedisplayshortskip}{5pt}
\setlength{\belowdisplayshortskip}{5pt}
\[
\mathbb{E}[R] = \sum_{j=1}^{|r|} r_j \cdot \mathrm{softmax}(r | d,i,a)_j,
\]%
}%
where $R$ is the random variable representing the rating, $r$ represents the rating values: $\{0,1\}$ for Question 1, $\{1,2,3,4,5\}$ for Question 2. $(d,i,a)$ represent the document, instruction, and answer.

Additionally,
we discuss a variant called \textit{Constrained Softmax ${n}$-shot},
where we contextualize the model with $n$ examples
(document-instruction-answer-rating tuples) in each of the prompts.

\paragraph{Self-Agreement}%
In this method, we test if the model is consistent with itself across rating generations by repeatedly sampling the rating from the LLM $n=7$ times. To diversify the samples, we experiment with various softmax temperatures, only to find that lower temperatures yield better results\footnote{\,Temperature is set to 0.1 for all reported Self-Agreement and Multi-LLM Agreement experiments.}.
The final rating is the arithmetic mean of the individual samples. We contextualize the model with $k=3$ examples in the prompt (see Figure~\ref{fig:self-agreement-prompt} in Appendix~\ref{app:prompts}). We also investigate the following variants:
\begin{itemize}[left=0pt,itemsep=-0.5em,topsep=0.25em,partopsep=0pt]
\item \textit{no intro}\quad Omitting the description of the task in the prompt and using only the $k$ examples.
\item \textit{rationale}\quad Asking the model to generate Chain of Thought-like ``rationales'' for the given rating to each few-shot example \citep{wei2023chainofthought}.
\item \textit{random}\quad Using the same hand-crafted examples (not occurring in the dataset) vs. picking $k$ random examples from the remaining documents in the dataset.
\end{itemize}

\begin{figure}[tb]
    \centering
    \includegraphics[trim=15pt 55pt 40pt 30pt,clip=true,width=0.48\textwidth]{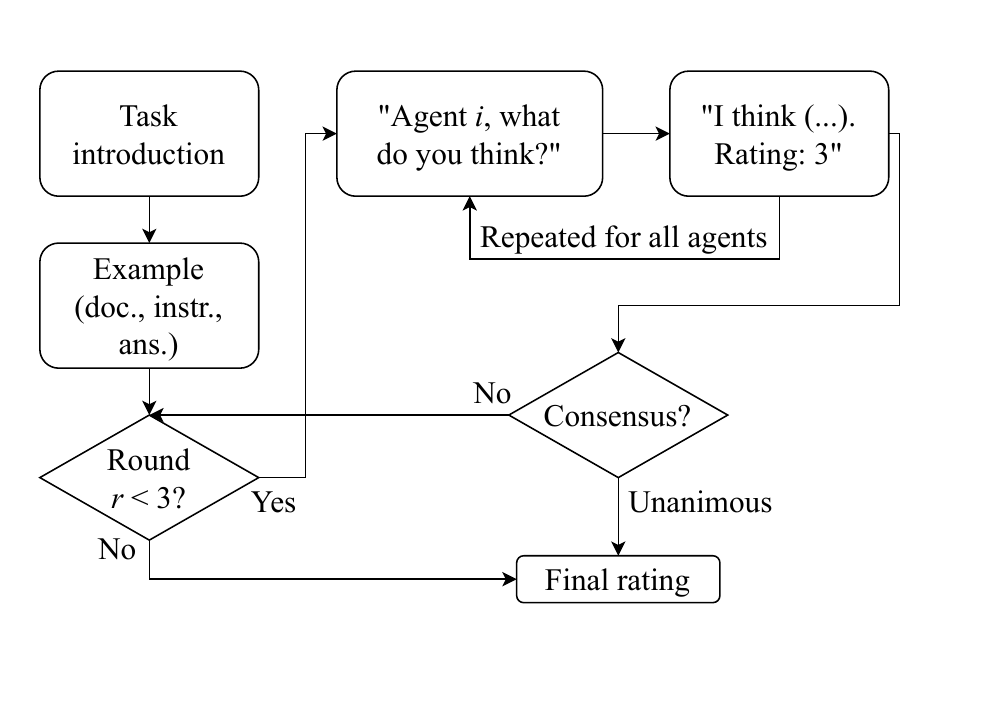}
    \caption{Multi-LLM agreement communication flow.}
    \label{fig:consensus-flowchart}
\end{figure}

\paragraph{Multi-LLM Agreement}
Recent works \cite{bakker2022finetuning,park2023generative} have used LLMs in conversational settings where all participant LLMs communicate with each other
and try to achieve a common goal. We propose a consensus-based metric where $k$ LLM instances\footnote{\,$k=3$ in all our experiments.}
debate amongst each other and try to arrive at a common assessment.
Though there are no restrictions on the LLMs to use, we evaluate the simplest case where each instance is the same LLM.
The rules of communication are set as follows (Figure~\ref{fig:consensus-flowchart}):

\begin{enumerate}[left=0pt,itemsep=-0.25em,topsep=0.25em,partopsep=0pt]
\item The models communicate amongst each other in a controlled manner for up to 3 rounds and try to arrive at a consensus. After at most 3 rounds, one of three outcomes occurs:
(i) \textit{unanimous agreement}: all 3 models agree. If this happens in the earlier rounds, the process ends immediately,
(ii) \textit{majority agreement}: one model disagrees with the other two, or
(iii) \textit{disagreement}: all 3 models disagree with each other.
\item In each round, all models provide a rating and a brief rationale.
The models do not have access to the other model outputs till the end of a round\footnote{\,Empirically, models tend to agree more easily
with each other when shown other models' ratings before the round ends.}.
Before the start of rounds two and three, they receive the ratings and rationales of all models from the previous rounds.
\end{enumerate}
The prompt for the models is presented in Figure~\ref{fig:consensus-prompt} of Appendix~\ref{app:prompts}.
This method is referred to as \textit{Multi-LLM Agreement} henceforth. We repeat the process $n=3$ times for added stability.

\section{Evaluating Agreement with Annotators}

As discussed in Section~\ref{sec:collection}, we asked annotators to provide a binary Yes/No rating answering whether a model output \textit{follows the instruction} and a qualitative rating from 1 to 5, representing \textit{how well} it follows the instruction.
Using meta-evaluation methods described below, we then study agreement between annotators and our evaluation methods.

\subsection{``Follows Instruction?''}\label{sec:roc}

For the binary rating, we compute a macro-averaged \textit{Area Under ROC Curve} (AUC ROC) statistic for each evaluation method. Using AUC ROC, we analyze the effectiveness of each method if they were used as binary classifiers for ``Does the output follow the instruction?'',
thereby assessing the degree to which they can replace human ratings. Since our classes are imbalanced towards ``Yes'' (Table~\ref{tab:model_comparison_answer}) we opt for the macro-averaged version of ROC AUC so that we can better detect which methods can accurately predict the ``No'' class.

\subsection{``How well?''}

\paragraph{Rank-based evaluation}
To analyze the ability of evaluation metrics to rank model outputs in relation to each other,
we compute \textit{Kendall's $\mathcal{T}_b$ rank distance} $d_{\mathcal{T}_b}$ among the model outputs for each document-instruction pair.
When the ranking produced by a metric is independent from human ranking, the value of $d_{\mathcal{T}_b}$ will be equal to $0.5$ in expectation.
Values below $0.5$ represent rankings that are similar to the human ranking,
values above $0.5$ represent orderings that are similar to the inverse of the human ranking.
As opposed to the $\mathcal{T}_b$ rank correlation coefficient, $d_{\mathcal{T}_b}$ has values in the range of $[0, 1]$ and can be interpreted as a distance function (lower is better):
$d_{\mathcal{T}_b} = (1-\mathcal{T}_b)\;/\;2.$
Compared to other forms of $\mathcal{T}$, $\mathcal{T}_b$ adjusts for ties: situations, where a metric or annotators give the same rating to two or more model outputs for one document-instruction pair.

For our human ratings, $\mathcal{T}_b$ is not defined for $9$ out of $300$ document-instruction pairs: the mean of the 3 annotators' ratings is constant for all 3 models, making it impossible to rank the models.
We report the mean and standard error of the rank distance $d_{\mathcal{T}_b}$ across all non-constant pairs.

\paragraph{Linear value correlation}
Additionally, we would like evaluation method outputs to align with annotators' notions of ``good'' or ``bad''. To study this, we compute Pearson's distance across all document-instruction-answer tuples:
$d_{|r|} = 1-|r|,$
where $r$ is Pearson's correlation coefficient between an evaluation method's values and the mean annotator rating.
Values of $d_{|r|}$ range from $0$ to $1$; the lower the value, the higher the linear correlation with human ratings.

\section{Results and Analysis}

\begin{table*}[tb]
\centering
\small
\begin{tabular}{l c cc}
\toprule
\multirow{2.5}{*}{Evaluation Method} & \multicolumn{1}{c}{Follows Instruction?} & \multicolumn{2}{c}{How Well?}\\
\cmidrule(lr){2-2}\cmidrule(lr){3-4}
& AUC ROC $\%\;\uparrow$ & $\overline{d_{\mathcal{T}_b}}\%\;\downarrow$ & $\overline{d_{|r|}}\%\;\downarrow$ \\
\midrule\multicolumn{4}{c}{\textbf{Reference-based baseline methods} (Section~\ref{sec:methods-ref})}\\ \midrule

\bleurt/\textsubscript{20} [references: GPT-4] & $\mathbf{78.5} \pm {1.9}$ & ${41.1} \pm {1.9}$ & ${50.9} \pm {2.9}$\\
\bleurt/\textsubscript{20} [references: \fpalmii{Lc}] & ${71.8} \pm {2.3}$ & ${48.8} \pm {1.9}$ & ${54.4} \pm {2.8}$\\[0.3em]

\rouge/\textsubscript{avg} [references: GPT-4] & $\mathbf{79.5} \pm {1.9}$ & $\mathbf{35.4} \pm {1.9}$ & ${52.6} \pm {2.8}$\\
\rouge/\textsubscript{avg} [references: \fpalmii{Lc}] & ${71.1} \pm {2.3}$ & ${46.7} \pm {1.9}$ & ${60.8} \pm {2.7}$\\

\midrule \multicolumn{4}{c}{\textbf{Reference-free baseline methods} (Section~\ref{sec:methods-free})}\\ \midrule

Sentence Count & ${39.5} \pm {3.1}$ & ${54.8} \pm {1.8}$ & ${72.7} \pm {2.3}$\\
Word Count & ${42.2} \pm {3.1}$ & ${51.4} \pm {2.0}$ & ${71.0} \pm {2.4}$\\[0.3em]

\bartscore/ \citep{bartscore} & ${68.4} \pm {2.5}$ & ${45.0} \pm {1.9}$ & ${74.7} \pm {2.2}$\\
\bartscore/\textsubscript{CNN} \citep{bartscore} & ${69.7} \pm {2.4}$ & ${43.7} \pm {1.9}$ & ${70.3} \pm {2.4}$\\
T5\textsubscript{ANLI} \citep{true} & ${71.9} \pm {2.3}$ & $\mathbf{38.8} \pm {1.9}$ & ${64.7} \pm {2.5}$\\

\midrule \multicolumn{4}{c}{\textbf{LLM-based reference-free methods} (Section~\ref{sec:methods-llm})}\\ \midrule

\palmii{S} Constrained Softmax & ${74.0} \pm {2.2}$ & ${43.6} \pm {1.9}$ & ${80.0} \pm {2.0}$\\
\palmii{L} Constrained Softmax & ${77.8} \pm {2.0}$ & ${39.9} \pm {1.9}$ & ${46.4} \pm {3.0}$\\[0.3em]

\fpalmii{Sc} Self-Agreement & ${67.2} \pm {2.5}$ & ${42.8} \pm {1.7}$ & ${56.7} \pm {2.7}$\\
\fpalmii{Lc} Self-Agreement & $\mathbf{81.7} \pm {1.7}$ & $\mathbf{37.1} \pm {1.7}$ & $\mathbf{39.5} \pm {3.1}$\\
\fpalmii{Lc} Self-Agreement (+ no intro) & $\mathbf{79.7} \pm {1.9}$ & $\mathbf{38.4} \pm {1.8}$ & ${45.7} \pm {3.0}$\\
\fpalmii{Lc} Self-Agreement (+ rationale) & ${75.0} \pm {2.1}$ & ${43.2} \pm {1.3}$ & ${50.5} \pm {2.9}$\\[0.3em]

\fpalmii{Sc} Self-Agreement (random) & ${69.0} \pm {2.4}$ & ${42.7} \pm {1.7}$ & ${58.3} \pm {2.7}$\\
\fpalmii{Lc} Self-Agreement (random) & $\mathbf{80.4} \pm {1.8}$ & $\mathbf{37.0} \pm {1.8}$ & $\mathbf{42.2} \pm {3.0}$\\
\fpalmii{Lc} Self-Agreement (random + no intro) & $\mathbf{78.2} \pm {1.9}$ & ${39.5} \pm {1.8}$ & ${50.2} \pm {2.9}$\\[0.3em]

\fpalmii{Sc} Multi-LLM Agreement & ${66.4} \pm {2.5}$ & ${45.7} \pm {1.2}$ & ${61.8} \pm {2.6}$\\
\fpalmii{Lc} Multi-LLM Agreement & ${67.1} \pm {2.5}$ & ${46.0} \pm {1.2}$ & ${58.7} \pm {2.7}$\\

\bottomrule
\end{tabular}
\caption{AUC ROC Curve measures how well methods predict Yes/No annotator responses on ``Follows Instruction?'' ($n=900$). For ``How Well?'' (1--5 rating), we report Kendall's rank distance $d_{\mathcal{T}_b}$ comparing evaluation methods' ranking of answers to that of annotators' ($n=291$) and Pearson's distance from mean annotator responses $d_{|r|}$ ($n=900$). All values are in \%, $\pm$ signifies standard error, $\uparrow$ signifies higher is better ($\downarrow$ lower is better). Methods highlighted in bold have overlapping confidence intervals with the best method per column. Non-deterministic methods (Self-Agreement, Multi-LLM Agreement) have been re-run $5\times$ and the mean is reported.}
\label{tab:results}
\end{table*}

We compare the effectiveness of evaluation methods on the three rating dimensions, based on the reported numbers for the binary rating ``Follows Instruction?'' and for the qualitative rating ``How well?'' in Table~\ref{tab:results}.
For both rating tasks, the two length-based heuristics perform the worst out of all methods, which suggests that the instructions are of good quality, as annotators are not strongly influenced by the length of model outputs.

\subsection{Predicting ``Follows Instruction?''}

First, we focus on how good of a binary classifier the methods are. We report the AUC ROC and its standard error (Section~\ref{sec:roc}) with respect to the human majority vote labels.

\paragraph{Reference-based methods} Having access to several reference answers that follow the instruction continues to be a good indicator when combined with \rouge/ or \bleurt/.
However, the results show that, when we have access to a capable LLM like \fpalmii{Lc}, it is better to use it directly as a reference-free evaluator, than sampling reference summaries from it and using reference-based metrics like \rouge/\textsubscript{avg} and \bleurt/\textsubscript{20}.

\paragraph{Reference-free methods} As expected, performance of each evaluation method improves
with model size.
We observe that standard error is usually higher ($>2.0$) when using \palmii{S} compared to \palmii{L} ($<2.0$), across different methods.
Combined with generally lower performance, methods using \palmii{S} as the underlying model are more noisy and produce less meaningful evaluations compared to methods using \palmii{L}. 

We also note that Multi-LLM Agreement approaches, while interesting, are outperformed by both Self-Agreement and Constrained Softmax approaches, irrespective of the model size.

For scoring-based approaches (Constrained Softmax), non-instruction-tuned LLMs outperform their instruction-tuned counterparts.
When generation is involved,
instruction-tuned models outperform their base versions.
This applies to rating generation, but also for generating answers directly.
We only report numbers of instruction-tuned LLMs for generation-based methods and correspondingly, only report numbers of non-instruction-tuned LLMs for scoring-based approaches.

\subsection{Predicting ``How Well?''}

In the case of qualitative ratings, obtaining a ranking of answers that matches the annotators' ranking proves to be difficult.
We note sensitivity in the analysis with respect to how ratings are aggregated per answer (majority vote or mean).
To minimize ties and maximize the use of annotator information, we use mean aggregation for the following analysis.

Observing $d_{\mathcal{T}_b}$ ranking performance,
\rouge/\textsubscript{avg} using GPT-4 model-generated answers seems to perform on-par with \fpalmii{Lc} Self-Agreement based methods,
as well as the 11B parameter T5\textsubscript{ANLI} model from \citet{true}.

\paragraph{Reference-based methods}  In our experiments, \bleurt/ performs worse than \rouge/ at relative ranking of model outputs.
Since \rouge/ is based on surface form, there is reason to believe that samples from different models in a single LM-family are closer in surface form than samples from different LM-families.
In Table~\ref{tab:agreement-analysis}, we analyze the effectiveness of methods at picking the best answers out of the 3 model outputs.
\textit{Perfect agreement} happens when the sets of annotator and metric ``winner'' models is equal.
\textit{Disagreement} occurs when the intersection between annotator and metric winners is empty.
Within disagreement, \textit{prefers own LM family} means the metric winners contained \textit{at least one} model output from the LM family the metric is based on.

We observe that when the evaluation model is sufficiently different from the rated models, the likelihood of evaluation models preferring their own LM family goes down.
However, when using a similar model, reference-based methods are more biased towards preferring their own LM family.
If human-written reference answers are unavailable, using a reference-free metric is preferable.

\begin{table}[tb]
\small
\centering
\setlength\tabcolsep{0.3em}
\begin{tabular}{@{} l@{}rrr@{}}
\toprule
Evaluation & Perfect & Disagree- & Prefers own\\
Method & agreement & ment & LM family\\\midrule
Constr. Softmax\textsubscript{\;\palmii{L}} & 35.7\%  & 40.3\% & 56.2\%\\\midrule
\rouge/\textsubscript{\;\fpalmii{Lc}} & 27.0\% & 48.0\% & 93.1\% \\
Constr. Softmax\textsubscript{\;\fpalmii{Lc}} & 23.0\% & 54.3\% & 71.8\%\\
Self-Agr.\textsubscript{\;\fpalmii{Lc}} & 25.7\% & 23.0\% & 72.5\% \\
\bottomrule
\end{tabular}
\caption{Agreement analysis with respect to mean qualitative ranking (``How well?'').}
\label{tab:agreement-analysis}
\end{table}

\paragraph{Reference-free methods}
Similarly to the binary rating, we observe that methods with larger underlying models perform better.
Likewise, reference-free methods based on \fpalmii{Lc} outperform their reference-based counterparts when using the same underlying model.
The base \palmii{L}
model with Constrained Softmax performs better and at lower cost than using the instruction-tuned \fpalmii{Lc} to generate reference summaries. With more available compute, one can further improve performance by leveraging multi-sampling Self-Agreement methods.

Interestingly, using \textit{random} examples in Self-Agreement decreases performance as opposed to hand-crafting a small ($k=4$) set of held-out examples. Contrary to intuition, using Chain-of-Thought approaches (\textit{rationale}) seems to degrade performance, but when removing the task description (\textit{no intro}) we do not observe a big drop.

When linear correlation $d_{|r|}$ with human ratings is required, methods that model the qualitative rating directly outperform more generic methods.

\section{Related Work}

\paragraph{Measuring instruction following with LLMs}
\citet{liu2023geval} use GPT-4 as a backbone model and study the correlation with human ratings on non-query-based summarization, finding a bias towards LLM-generated text. We do not study this aspect, as our rating task focuses on model-generated text. \citet{fu2023gptscore} propose a zero-shot approach for multi-faceted evaluation of text generation.

An increase in interest for improving instruction-following capabilities of LLMs has resulted in the creation of multiple datasets. FLAN \cite{weifinetuned} and Natural Instructions \cite{naturalinstructions} were two of the earlier datasets which turned standard NLP tasks (e.g.~sentiment classification, question-answering) into instruction following tasks. Other works like Self-Instruct \cite{selfinstruct}, Super-NaturalInstructions \citep{supernatural}, and the H4 instruction dataset \cite{h4} curate human-written instruction and answer pairs. \citet{guo-etal-2023-hc3} and \citet{alpaca-cot} collect instruction-answer pairs from LLM generations. All of them use standard NLP metrics or human annotation to evaluate the model outputs.

\paragraph{Model-based metrics}
A large body of prior work focuses on model-based approaches fine-tuned on human ratings.
Usually, encoder models such as \textsc{BERTScore} \cite{zhangbertscore} or \bleurt/ \cite{sellam2020bleurt} are used, but encoder-decoder methods exist as well (\bartscore/, \citealp{bartscore}). We focus on low-resource zero/few-shot methods using larger, decoder-containing models from PaLM and GPT families.

\paragraph{Human evaluation}
\citet{kryscinski2018improving,huang-etal-2020-achieved,shen2022mred} and several others have resorted to human evaluation for analyzing the quality of reference summaries and model outputs. They adopt a Likert-type scale for rating individual aspects of generated text. \citet{fan2018hierarchical,fabbri-etal-2019-multi,shen2022sentbs} and others perform side-by-side comparisons of two or more model-generated summaries and use Elo, or other rating systems to build rankings of models.

\paragraph{LLM evaluation}
Many recent works use LLMs as evaluators for summarization tasks. \citet{wu2023large} use LLMs with ``different persona'' to evaluate summaries from various perspectives.
\citet{luo2023chatgpt} examine if LLMs can be used to detect factual inconsistencies.
Concurrent to our work, \citet{liu2023revisiting} curate a human-evaluation dataset consisting of 22,000 summary-level annotations
and perform a study of various automatic and LLM-based metrics for summarization and call for more rigorous evaluation of LLM performance.

\section{Conclusion}
In this work, we investigate the effectiveness of multiple evaluation methods in quantifying the degree to which LLM-generated text follows user-given instructions.
We release \dataset/, a new short-form dataset of $300$ document-instruction pairs with $3$ answers each. All of the $900$ answers are rated by at least $3$ human annotators.
When analyzing agreement between evaluation methods and human judgment,
we find that established metrics, such as \rouge/ and \bleurt/ are not effective at quantifying LLMs' instruction-following ability.
LLM-based evaluation methods tend to have stronger correlation with annotator judgment, without requiring high-quality reference answers.
We hope that the introduced evaluation framework is adopted by the community for evaluating instruction-following abilities of LLMs, possibly expanding into more tasks, domains, and examples.

\section*{Acknowledgements}
The authors would like to thank Lucas Werner for building the human annotation tool which dramatically simplified the creation of the \dataset/ dataset, as well as all the 14 annotators and their managers for their tireless and accurate work.
Thank you also to Hassan Mansoor and Misha Khalman, who have helped with exploratory work on evaluation methods and early data collection.
We also thank the anonymous reviewers for their constructive feedback.
Lastly, we would like to thank Matt Sharifi, Jindong Chen and and Blaise Ag\"{u}era y Arcas for their feedback on the paper draft.

\bibliography{emnlp2023}
\bibliographystyle{acl_natbib}

\clearpage
\appendix

\section{Limitations}
While the presented data offers a variety (e.g.~diverse origin texts), a drawback to our work is that we only consider the task of instruction-based summarization (e.g.~long-form question answering, query-driven summarization, stylistic summarization) as such. The extent to which  metrics generalize to other tasks is not yet explored. Furthermore, for language diversity, the proposed benchmarks are restricted to English only.  However, we hope that this initial benchmark allows further work to consider a larger range of tasks as well as exploration for how these benchmarks generalize to other languages.

Our correlation with human judgment analysis on the qualitative rating (``How Well?'') has a limitation where the annotators do not provide sufficient signal to distinguish between the 3 answers. This happens in only 9 out of the 300 document-instruction pairs and we chose to skip those pairs in the analysis for this rating task. The motivation for doing this is that our focus is on the cases where there is sufficient signal from the human annotators when an answer is better than another.

We acknowledge that relying on human ratings as a ground truth has drawbacks, especially as summarization is notoriously difficult to evaluate due to the subjective nature. To mitigate this, we provide extensive training and feedback to annotators and are in active communication throughout the annotation process to provide clarifications. The annotators used in our experiment have over a year of experience with rating NLU tasks. However, a limitation is that our annotator pool represents individuals from similar backgrounds, which may mean other populations would have differing quality perspectives. The background statistics of annotators can be found in Appendix~\ref{app:demographics}.

\section{Ethics Statement}
The alignment of model behavior with user expectations is a crucial area of research, and we recognize the importance of contributing to the development of benchmarking methods for instruction following. Our work represents a step towards benchmarking how LLMs can self-evaluate their performance in the task of summarization. However, there are still many other aspects of summary quality, such as factuality, that warrant further exploration due to their significant downstream implications.

A model's ability to follow instructions for a specific task, such as summarization, may not reflect the overall proficiency in instruction following. As such, these metrics serve as proxies to estimate the extent to which task instructions are adhered to within the context of summarization. Given the ongoing discussions regarding the risks associated with LLMs, this distinction is relevant.

During dataset construction, it is important to acknowledge the ethical concerns arising from the use of publicly sourced data without explicit permission from the original parties. While the data we employ is derived from previously released datasets, the examples are generated using LLMs trained on large, uncurated, static datasets obtained from the internet.

\section{Annotator methodology}\label{app:annotators}

\begin{figure}[tb]
    \hspace{-10px}\input{rater_question_distrib_histogram.pgf}
    \caption{Number of questions annotated by each human annotator. Annotator IDs pseudonymized to capital letters.}
    \label{fig:rater-hist}
\end{figure}
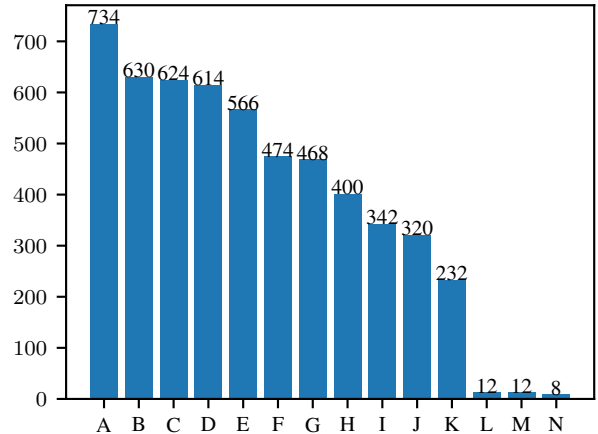

\subsection{Annotation UI}

\begin{figure*}[tb]
    \centering
    \includegraphics[width=0.8\textwidth]{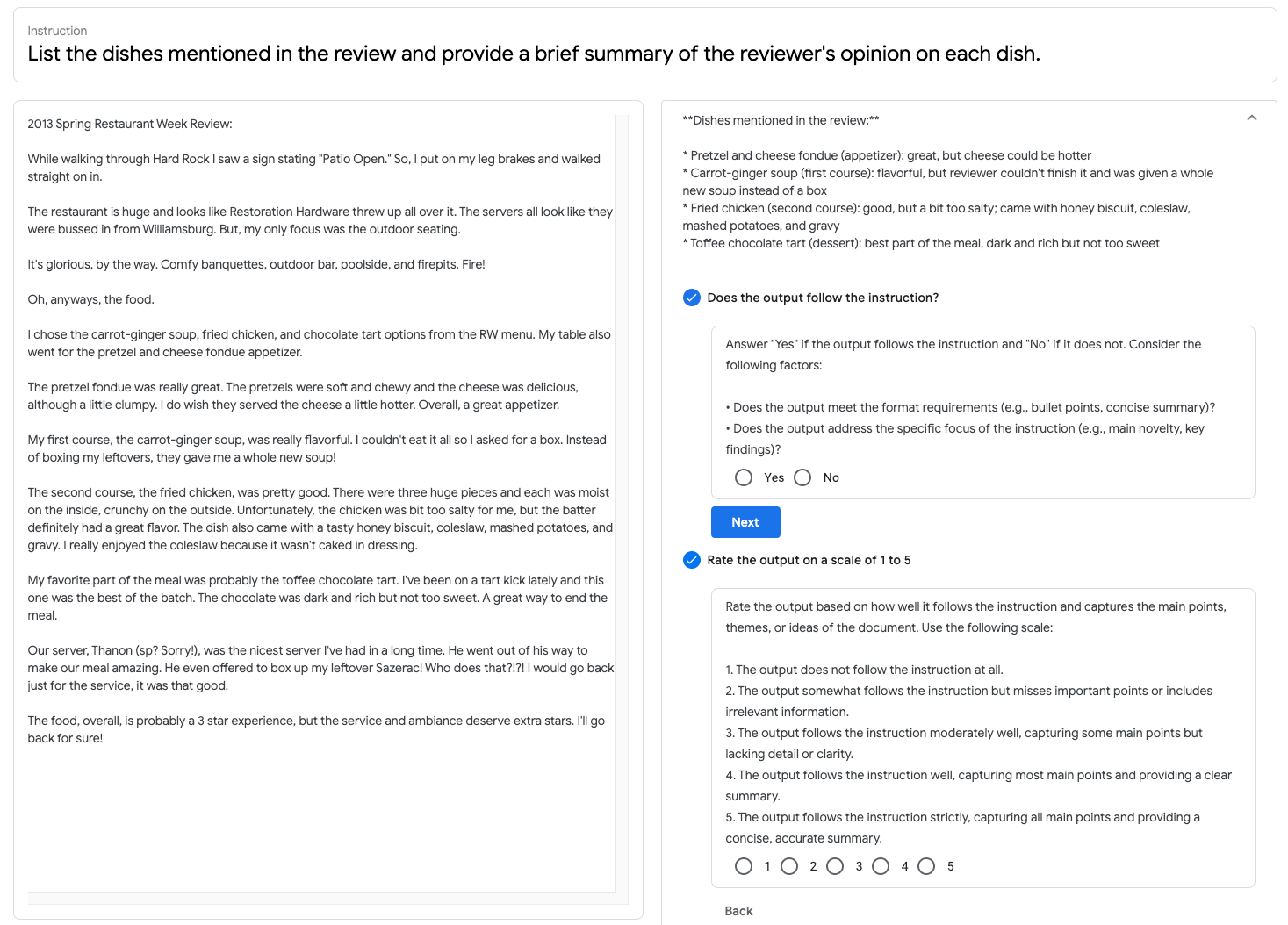}
    \caption{Example screenshot of the annotator UI.}
    \label{fig:rater-ui}
\end{figure*}

In Figure~\ref{fig:rater-ui} we illustrate the user interface used for collecting the dataset. Annotators follow a multi-step process, by first answering ``Does the output follow the instruction?'' followed by ``Rate the output on a scale of 1 to 5'' to qualitatively assess the answer.

The UI also allows annotators to navigate through the provided content and highlight words that appear either in the answer or in the original text. Annotators can use this as a way to verify that content is present in both the output and input.

\subsection{Annotator demographics}\label{app:demographics}

Table~\ref{tab:participants} presents the results of an optional questionnaire given to our annotators, aimed at understanding their background factors. Out of a total of 14 annotators, we have received responses from 7 individuals, who collectively accounted for approximately $65\%$ of the annotation coverage for our dataset (Figure~\ref{fig:rater-hist}). This information allows us to gain a better understanding of the perspectives and experiences of our annotators, which can impact the annotation outcomes.

\begin{table*}[tb]
\centering
\begin{tabular}{@{} lc|lc|lc|lc @{}} \toprule
\multicolumn{2}{c}{Proficiency} & \multicolumn{2}{c}{Education} & \multicolumn{2}{c}{Age range} & \multicolumn{2}{c}{\vtop{\hbox{\strut Hours reading}\hbox{\strut English per week}}} \\ \midrule
Native & $1/7$ & Graduate  & $6/7$    & $18$--$24$ & $3/7$ & $0$--$5$      & $1/7$  \\
Near native   & $1/7$  & Undergraduate & $1/7$     & $25$--$34$ & $4/7$   & $5$--$10$     & $2/7$    \\
Advanced   & $5/7$  & High School   & $0/7$     & $35$--$44$ & $0/7$ & $10$--$15$    & $1/7$    \\
Intermediate  & $0/7$  & Vocational Training     & $0/7$    & $45$--$54$ & $0/7$   & $15$--$20$    & $0/7$    \\
Beginner   & $0/7$  & No formal education     & $0/7$       & $55+$ & $0/7$ & $20+$   & $3/7$    \\ \bottomrule
\end{tabular}
\caption{Background statistics for annotators.}
\label{tab:participants}
\end{table*}

\section{Annotator Guidelines}\label{app:guidelines}
\subsection{Objective}
The goal of this task is to evaluate the quality of summaries generated based on given instructions. You will be provided with a document, an instruction, and an output (summary). Your task is to answer two questions:
\begin{enumerate}
    \item Does the output follow the instruction? (Yes/No), and
    \item Rate the output on a scale of 1 to 5, with 1 indicating that the output does not follow the instruction at all, and 5 indicating that the output follows the instruction strictly.
\end{enumerate}

\subsection{General Guidelines}
\paragraph{Understanding the Document}
Before evaluating the output, make sure you have a clear understanding of the document. The document can be a news article, a chat conversation, an email, etc. Read the document carefully and identify the main points, themes, or ideas.

\paragraph{Analyzing the Instruction}
The instruction will be related to summarization. It can be general (e.g.~``Summarize in 3 bullet points'') or specific to the paragraph (e.g.~``Summarize the main novelty of the research work concisely''). Make sure you understand the instruction and its requirements.

\paragraph{Evaluating the Output}
Compare the output with the document and the instruction. Check if the output follows the instruction and captures the main points, themes, or ideas of the document.

\paragraph{Evaluation Criteria}
For Question 1, answer ``Yes'' if the output follows the instruction and ``No'' if it does not. Consider the following factors: (i) does the output meet the format requirements (e.g.~bullet points, concise summary)? and (ii) does the output address the specific focus of the instruction (e.g.~main novelty, key findings)?

For Question 2, rate the output based on how well it follows the instruction and captures the main points, themes, or ideas of the document. Use the following scale:
\begin{enumerate}[left=0pt,itemsep=0pt,topsep=0.5em,partopsep=0pt]
\item The output does not follow the instruction at all.
\item The output somewhat follows the instruction but misses important points or includes irrelevant information.
\item The output follows the instruction moderately well, capturing some main points but lacking detail or clarity.
\item The output follows the instruction well, capturing most main points and providing a clear summary.
\item The output follows the instruction strictly, capturing all main points and providing a concise, accurate summary.
\end{enumerate}

\subsection{FAQs}
\paragraph{What if the output is well-written but does not follow the instruction?}
Rate the output based on how well it follows the instruction, not on its writing quality. If the output does not follow the instruction, give it a low rating.

\paragraph{What if the output follows the instruction but has grammatical errors or typos?}
Focus on the content and adherence to the instruction. Minor grammatical errors or typos should not significantly impact the rating unless they affect the clarity or accuracy of the summary.

\paragraph{What if the output is too long or too short?}
Consider whether the output meets the requirements of the instruction. If the instruction specifies a length (e.g.~``Summarize in 3 bullet points''), the output should adhere to that length. If the output is too long or too short, it may not follow the instructions strictly, and you should adjust the rating accordingly.

\paragraph{What if the output is accurate but not concise?}
If the instruction requires a concise summary, the output should be brief and to the point. If the output is accurate but not concise, it may not follow the instructions strictly, and you should adjust the rating accordingly.

\section{Prompts}\label{app:prompts}

List of prompts used in different parts of the paper:
\begin{itemize}[left=0pt,itemsep=0pt,topsep=0.5em,partopsep=0pt]
    \item GPT-4 prompt for generating \dataset/ instructions: Figure~\ref{fig:gpt4-instr-creation-prompt}.
    \item Self-agreement prompt: Figure~\ref{fig:self-agreement-prompt}.
    \item Multi-LLM agreement prompt: Figure~\ref{fig:consensus-prompt}.
    \item Constrained Softmax prompt: Figure~\ref{fig:scoring-prompt}.
\end{itemize}

\begin{figure*}[tb]
    \centering
    \small
    \begin{boxed2}
Read the paragraph given by the user and generate a list of 3-5 instructions for human annotators. Each instruction must be in a new line. \\ \\
The instructions must be related to the task of summarization. Some general examples are:
Summarize in 3 bullet points. \\
Write the main topics of the document in 2 sentences. \\
Summarize the paragraph in not more than 20 words. \\ \\
However, you can ask them to perform something specific related to the content of the paragraph. \\
Summarize the main novelty of the research work concisely. \\
Summarize the cleaning tips using soap and sponge in details for me so I sound like a professional. \\
Summarize the purpose of the dialogue and then convert each person's opinion into a bullet list while keeping their orders. \\ \\
Be as creative as possible, and use the information present in the paragraph to make the instructions unique.
    \end{boxed2}
    \caption{Prompt given to GPT-4 for creating the instructions.}
    \label{fig:gpt4-instr-creation-prompt}
\end{figure*}

\begin{figure*}[tb]
\centering
\small
\begin{boxed2}
You are given a document, an instruction, and a candidate answer.\\
You have to evaluate the answer based on how well it follows the instructions on a scale of 1 to 5 (larger is better), and provide a rationale.\\
Carefully evaluate the various constraints that may be present in the instructions.\\
\\
----\\
\\
Document:\\
\{document\}\\
\\
Instruction:\\
\{instruction\}\\
\\
Answer:\\
\{answer\}\\
\\
Rating:%
\end{boxed2}
\caption{Self-agreement prompt. The bottom part under and including ``----'' is repeated for $n>1$-shot variants.}
\label{fig:self-agreement-prompt}
\end{figure*}

\begin{figure*}[tb]
    \centering
    \small
    \begin{boxed2}
This is a chat room with AI assistants that specialize in summarizing and question answering.\\
You are given a paragraph of text, an instruction, and a candidate answer.\\
You have to evaluate the answer based on how well it follows the instructions on a scale of 1 to 5.\\
Carefully evaluate the various constraints that may be present in the instruction.\\
After evaluation, present a brief rationale not exceeding 2-3 sentences, and your rating, to the AI assistants.\\
If there is consensus among the AI assistants, the rating will be accepted.\\
If there is no consensus, you should read the rationale of the other AI assistants and try to reach a consensus by either changing your rating or convincing the other assistants to change theirs.\\
You will be given 3 chances to reach a consensus.\\
Always try to reach a consensus.\\
Remember, end your response with 'Rating:'.\\
\\
Document:\\
\{document\}\\
\\
Instruction:\\
\{instruction\}\\
\\
Answer:\\
\{answer\}\\
\\
(User: Agent \{aid\}, please share your response.)\\
Agent \{aid\}: ... rationale ... Rating: 4.\\
...%
\end{boxed2}
\caption{Prompt given to the models before the consensus discussion.}
\label{fig:consensus-prompt}
\end{figure*}

\begin{figure*}[tb]
\centering
\small
Question \#1
\begin{boxed2}
Does the output follow the instruction? Rate ``Yes'' if the output follows the instruction and ``No'' if it does not. Consider the following factors:\\
* Does the output meet the format requirements (e.g., bullet points, concise summary)?\\
* Does the output address the specific focus of the instruction (e.g., main novelty, key findings)?\\
\\
Document:\\
\{document\}\\
\\
Instruction:\\
\{instruction\}\\
\\
Output:\\
\{answer\}\\
\\
Rating:%
\end{boxed2}

Question \#2
\begin{boxed2}
Rate the output on a scale of 1 to 5. Rate the output based on how well it follows the instruction and captures the main points, themes, or ideas of the document. Use the following scale:\\
1. The output does not follow the instruction at all.\\
2. The output somewhat follows the instruction but misses important points or includes irrelevant information.\\
3. The output follows the instruction moderately well, capturing some main points but lacking detail or clarity.\\
4. The output follows the instruction well, capturing most main points and providing a clear summary.\\
5. The output follows the instruction strictly, capturing all main points and providing a concise, accurate summary.\\
\\
Document:\\
\{document\}\\
\\
Instruction:\\
\{instruction\}\\
\\
Output:\\
\{answer\}\\
\\
Rating:%
\end{boxed2}
    \caption{Prompts for Constrained Softmax-based methods.}
    \label{fig:scoring-prompt}
\end{figure*}

\end{document}

%% file: rater_local_alpha_histogram.pgf
%% Creator: Matplotlib, PGF backend
%%
%% To include the figure in your LaTeX document, write
%%   \input{<filename>.pgf}
%%
%% Make sure the required packages are loaded in your preamble
%%   \usepackage{pgf}
%%
%% Also ensure that all the required font packages are loaded; for instance,
%% the lmodern package is sometimes necessary when using math font.
%%   \usepackage{lmodern}
%%
%% Figures using additional raster images can only be included by \input if
%% they are in the same directory as the main LaTeX file. For loading figures
%% from other directories you can use the `import` package
%%   \usepackage{import}
%%
%% and then include the figures with
%%   \import{<path to file>}{<filename>.pgf}
%%
%% Matplotlib used the following preamble
%%   
%%   \makeatletter\@ifpackageloaded{underscore}{}{\usepackage[strings]{underscore}}\makeatother
%%
\begingroup%
\makeatletter%
\begin{pgfpicture}%
\pgfpathrectangle{\pgfpointorigin}{\pgfqpoint{3.100000in}{2.250000in}}%
\pgfusepath{use as bounding box, clip}%
\begin{pgfscope}%
\pgfsetbuttcap%
\pgfsetmiterjoin%
\pgfsetlinewidth{0.000000pt}%
\definecolor{currentstroke}{rgb}{1.000000,1.000000,1.000000}%
\pgfsetstrokecolor{currentstroke}%
\pgfsetstrokeopacity{0.000000}%
\pgfsetdash{}{0pt}%
\pgfpathmoveto{\pgfqpoint{0.000000in}{0.000000in}}%
\pgfpathlineto{\pgfqpoint{3.100000in}{0.000000in}}%
\pgfpathlineto{\pgfqpoint{3.100000in}{2.250000in}}%
\pgfpathlineto{\pgfqpoint{0.000000in}{2.250000in}}%
\pgfpathlineto{\pgfqpoint{0.000000in}{0.000000in}}%
\pgfpathclose%
\pgfusepath{}%
\end{pgfscope}%
\begin{pgfscope}%
\pgfsetbuttcap%
\pgfsetmiterjoin%
\definecolor{currentfill}{rgb}{1.000000,1.000000,1.000000}%
\pgfsetfillcolor{currentfill}%
\pgfsetlinewidth{0.000000pt}%
\definecolor{currentstroke}{rgb}{0.000000,0.000000,0.000000}%
\pgfsetstrokecolor{currentstroke}%
\pgfsetstrokeopacity{0.000000}%
\pgfsetdash{}{0pt}%
\pgfpathmoveto{\pgfqpoint{0.394308in}{0.315988in}}%
\pgfpathlineto{\pgfqpoint{2.980000in}{0.315988in}}%
\pgfpathlineto{\pgfqpoint{2.980000in}{2.130000in}}%
\pgfpathlineto{\pgfqpoint{0.394308in}{2.130000in}}%
\pgfpathlineto{\pgfqpoint{0.394308in}{0.315988in}}%
\pgfpathclose%
\pgfusepath{fill}%
\end{pgfscope}%
\begin{pgfscope}%
\pgfpathrectangle{\pgfqpoint{0.394308in}{0.315988in}}{\pgfqpoint{2.585692in}{1.814012in}}%
\pgfusepath{clip}%
\pgfsetbuttcap%
\pgfsetmiterjoin%
\definecolor{currentfill}{rgb}{0.121569,0.466667,0.705882}%
\pgfsetfillcolor{currentfill}%
\pgfsetfillopacity{0.500000}%
\pgfsetlinewidth{0.000000pt}%
\definecolor{currentstroke}{rgb}{0.000000,0.000000,0.000000}%
\pgfsetstrokecolor{currentstroke}%
\pgfsetstrokeopacity{0.500000}%
\pgfsetdash{}{0pt}%
\pgfpathmoveto{\pgfqpoint{0.511840in}{0.315988in}}%
\pgfpathlineto{\pgfqpoint{0.658754in}{0.315988in}}%
\pgfpathlineto{\pgfqpoint{0.658754in}{0.315988in}}%
\pgfpathlineto{\pgfqpoint{0.511840in}{0.315988in}}%
\pgfpathlineto{\pgfqpoint{0.511840in}{0.315988in}}%
\pgfpathclose%
\pgfusepath{fill}%
\end{pgfscope}%
\begin{pgfscope}%
\pgfpathrectangle{\pgfqpoint{0.394308in}{0.315988in}}{\pgfqpoint{2.585692in}{1.814012in}}%
\pgfusepath{clip}%
\pgfsetbuttcap%
\pgfsetmiterjoin%
\definecolor{currentfill}{rgb}{0.121569,0.466667,0.705882}%
\pgfsetfillcolor{currentfill}%
\pgfsetfillopacity{0.500000}%
\pgfsetlinewidth{0.000000pt}%
\definecolor{currentstroke}{rgb}{0.000000,0.000000,0.000000}%
\pgfsetstrokecolor{currentstroke}%
\pgfsetstrokeopacity{0.500000}%
\pgfsetdash{}{0pt}%
\pgfpathmoveto{\pgfqpoint{0.658754in}{0.315988in}}%
\pgfpathlineto{\pgfqpoint{0.805668in}{0.315988in}}%
\pgfpathlineto{\pgfqpoint{0.805668in}{0.315988in}}%
\pgfpathlineto{\pgfqpoint{0.658754in}{0.315988in}}%
\pgfpathlineto{\pgfqpoint{0.658754in}{0.315988in}}%
\pgfpathclose%
\pgfusepath{fill}%
\end{pgfscope}%
\begin{pgfscope}%
\pgfpathrectangle{\pgfqpoint{0.394308in}{0.315988in}}{\pgfqpoint{2.585692in}{1.814012in}}%
\pgfusepath{clip}%
\pgfsetbuttcap%
\pgfsetmiterjoin%
\definecolor{currentfill}{rgb}{0.121569,0.466667,0.705882}%
\pgfsetfillcolor{currentfill}%
\pgfsetfillopacity{0.500000}%
\pgfsetlinewidth{0.000000pt}%
\definecolor{currentstroke}{rgb}{0.000000,0.000000,0.000000}%
\pgfsetstrokecolor{currentstroke}%
\pgfsetstrokeopacity{0.500000}%
\pgfsetdash{}{0pt}%
\pgfpathmoveto{\pgfqpoint{0.805668in}{0.315988in}}%
\pgfpathlineto{\pgfqpoint{0.952582in}{0.315988in}}%
\pgfpathlineto{\pgfqpoint{0.952582in}{0.315988in}}%
\pgfpathlineto{\pgfqpoint{0.805668in}{0.315988in}}%
\pgfpathlineto{\pgfqpoint{0.805668in}{0.315988in}}%
\pgfpathclose%
\pgfusepath{fill}%
\end{pgfscope}%
\begin{pgfscope}%
\pgfpathrectangle{\pgfqpoint{0.394308in}{0.315988in}}{\pgfqpoint{2.585692in}{1.814012in}}%
\pgfusepath{clip}%
\pgfsetbuttcap%
\pgfsetmiterjoin%
\definecolor{currentfill}{rgb}{0.121569,0.466667,0.705882}%
\pgfsetfillcolor{currentfill}%
\pgfsetfillopacity{0.500000}%
\pgfsetlinewidth{0.000000pt}%
\definecolor{currentstroke}{rgb}{0.000000,0.000000,0.000000}%
\pgfsetstrokecolor{currentstroke}%
\pgfsetstrokeopacity{0.500000}%
\pgfsetdash{}{0pt}%
\pgfpathmoveto{\pgfqpoint{0.952582in}{0.315988in}}%
\pgfpathlineto{\pgfqpoint{1.099497in}{0.315988in}}%
\pgfpathlineto{\pgfqpoint{1.099497in}{0.315988in}}%
\pgfpathlineto{\pgfqpoint{0.952582in}{0.315988in}}%
\pgfpathlineto{\pgfqpoint{0.952582in}{0.315988in}}%
\pgfpathclose%
\pgfusepath{fill}%
\end{pgfscope}%
\begin{pgfscope}%
\pgfpathrectangle{\pgfqpoint{0.394308in}{0.315988in}}{\pgfqpoint{2.585692in}{1.814012in}}%
\pgfusepath{clip}%
\pgfsetbuttcap%
\pgfsetmiterjoin%
\definecolor{currentfill}{rgb}{0.121569,0.466667,0.705882}%
\pgfsetfillcolor{currentfill}%
\pgfsetfillopacity{0.500000}%
\pgfsetlinewidth{0.000000pt}%
\definecolor{currentstroke}{rgb}{0.000000,0.000000,0.000000}%
\pgfsetstrokecolor{currentstroke}%
\pgfsetstrokeopacity{0.500000}%
\pgfsetdash{}{0pt}%
\pgfpathmoveto{\pgfqpoint{1.099497in}{0.315988in}}%
\pgfpathlineto{\pgfqpoint{1.246411in}{0.315988in}}%
\pgfpathlineto{\pgfqpoint{1.246411in}{0.315988in}}%
\pgfpathlineto{\pgfqpoint{1.099497in}{0.315988in}}%
\pgfpathlineto{\pgfqpoint{1.099497in}{0.315988in}}%
\pgfpathclose%
\pgfusepath{fill}%
\end{pgfscope}%
\begin{pgfscope}%
\pgfpathrectangle{\pgfqpoint{0.394308in}{0.315988in}}{\pgfqpoint{2.585692in}{1.814012in}}%
\pgfusepath{clip}%
\pgfsetbuttcap%
\pgfsetmiterjoin%
\definecolor{currentfill}{rgb}{0.121569,0.466667,0.705882}%
\pgfsetfillcolor{currentfill}%
\pgfsetfillopacity{0.500000}%
\pgfsetlinewidth{0.000000pt}%
\definecolor{currentstroke}{rgb}{0.000000,0.000000,0.000000}%
\pgfsetstrokecolor{currentstroke}%
\pgfsetstrokeopacity{0.500000}%
\pgfsetdash{}{0pt}%
\pgfpathmoveto{\pgfqpoint{1.246411in}{0.315988in}}%
\pgfpathlineto{\pgfqpoint{1.393325in}{0.315988in}}%
\pgfpathlineto{\pgfqpoint{1.393325in}{0.315988in}}%
\pgfpathlineto{\pgfqpoint{1.246411in}{0.315988in}}%
\pgfpathlineto{\pgfqpoint{1.246411in}{0.315988in}}%
\pgfpathclose%
\pgfusepath{fill}%
\end{pgfscope}%
\begin{pgfscope}%
\pgfpathrectangle{\pgfqpoint{0.394308in}{0.315988in}}{\pgfqpoint{2.585692in}{1.814012in}}%
\pgfusepath{clip}%
\pgfsetbuttcap%
\pgfsetmiterjoin%
\definecolor{currentfill}{rgb}{0.121569,0.466667,0.705882}%
\pgfsetfillcolor{currentfill}%
\pgfsetfillopacity{0.500000}%
\pgfsetlinewidth{0.000000pt}%
\definecolor{currentstroke}{rgb}{0.000000,0.000000,0.000000}%
\pgfsetstrokecolor{currentstroke}%
\pgfsetstrokeopacity{0.500000}%
\pgfsetdash{}{0pt}%
\pgfpathmoveto{\pgfqpoint{1.393325in}{0.315988in}}%
\pgfpathlineto{\pgfqpoint{1.540240in}{0.315988in}}%
\pgfpathlineto{\pgfqpoint{1.540240in}{0.315988in}}%
\pgfpathlineto{\pgfqpoint{1.393325in}{0.315988in}}%
\pgfpathlineto{\pgfqpoint{1.393325in}{0.315988in}}%
\pgfpathclose%
\pgfusepath{fill}%
\end{pgfscope}%
\begin{pgfscope}%
\pgfpathrectangle{\pgfqpoint{0.394308in}{0.315988in}}{\pgfqpoint{2.585692in}{1.814012in}}%
\pgfusepath{clip}%
\pgfsetbuttcap%
\pgfsetmiterjoin%
\definecolor{currentfill}{rgb}{0.121569,0.466667,0.705882}%
\pgfsetfillcolor{currentfill}%
\pgfsetfillopacity{0.500000}%
\pgfsetlinewidth{0.000000pt}%
\definecolor{currentstroke}{rgb}{0.000000,0.000000,0.000000}%
\pgfsetstrokecolor{currentstroke}%
\pgfsetstrokeopacity{0.500000}%
\pgfsetdash{}{0pt}%
\pgfpathmoveto{\pgfqpoint{1.540240in}{0.315988in}}%
\pgfpathlineto{\pgfqpoint{1.687154in}{0.315988in}}%
\pgfpathlineto{\pgfqpoint{1.687154in}{0.315988in}}%
\pgfpathlineto{\pgfqpoint{1.540240in}{0.315988in}}%
\pgfpathlineto{\pgfqpoint{1.540240in}{0.315988in}}%
\pgfpathclose%
\pgfusepath{fill}%
\end{pgfscope}%
\begin{pgfscope}%
\pgfpathrectangle{\pgfqpoint{0.394308in}{0.315988in}}{\pgfqpoint{2.585692in}{1.814012in}}%
\pgfusepath{clip}%
\pgfsetbuttcap%
\pgfsetmiterjoin%
\definecolor{currentfill}{rgb}{0.121569,0.466667,0.705882}%
\pgfsetfillcolor{currentfill}%
\pgfsetfillopacity{0.500000}%
\pgfsetlinewidth{0.000000pt}%
\definecolor{currentstroke}{rgb}{0.000000,0.000000,0.000000}%
\pgfsetstrokecolor{currentstroke}%
\pgfsetstrokeopacity{0.500000}%
\pgfsetdash{}{0pt}%
\pgfpathmoveto{\pgfqpoint{1.687154in}{0.315988in}}%
\pgfpathlineto{\pgfqpoint{1.834068in}{0.315988in}}%
\pgfpathlineto{\pgfqpoint{1.834068in}{0.463574in}}%
\pgfpathlineto{\pgfqpoint{1.687154in}{0.463574in}}%
\pgfpathlineto{\pgfqpoint{1.687154in}{0.315988in}}%
\pgfpathclose%
\pgfusepath{fill}%
\end{pgfscope}%
\begin{pgfscope}%
\pgfpathrectangle{\pgfqpoint{0.394308in}{0.315988in}}{\pgfqpoint{2.585692in}{1.814012in}}%
\pgfusepath{clip}%
\pgfsetbuttcap%
\pgfsetmiterjoin%
\definecolor{currentfill}{rgb}{0.121569,0.466667,0.705882}%
\pgfsetfillcolor{currentfill}%
\pgfsetfillopacity{0.500000}%
\pgfsetlinewidth{0.000000pt}%
\definecolor{currentstroke}{rgb}{0.000000,0.000000,0.000000}%
\pgfsetstrokecolor{currentstroke}%
\pgfsetstrokeopacity{0.500000}%
\pgfsetdash{}{0pt}%
\pgfpathmoveto{\pgfqpoint{1.834068in}{0.315988in}}%
\pgfpathlineto{\pgfqpoint{1.980983in}{0.315988in}}%
\pgfpathlineto{\pgfqpoint{1.980983in}{0.376759in}}%
\pgfpathlineto{\pgfqpoint{1.834068in}{0.376759in}}%
\pgfpathlineto{\pgfqpoint{1.834068in}{0.315988in}}%
\pgfpathclose%
\pgfusepath{fill}%
\end{pgfscope}%
\begin{pgfscope}%
\pgfpathrectangle{\pgfqpoint{0.394308in}{0.315988in}}{\pgfqpoint{2.585692in}{1.814012in}}%
\pgfusepath{clip}%
\pgfsetbuttcap%
\pgfsetmiterjoin%
\definecolor{currentfill}{rgb}{0.121569,0.466667,0.705882}%
\pgfsetfillcolor{currentfill}%
\pgfsetfillopacity{0.500000}%
\pgfsetlinewidth{0.000000pt}%
\definecolor{currentstroke}{rgb}{0.000000,0.000000,0.000000}%
\pgfsetstrokecolor{currentstroke}%
\pgfsetstrokeopacity{0.500000}%
\pgfsetdash{}{0pt}%
\pgfpathmoveto{\pgfqpoint{1.980983in}{0.315988in}}%
\pgfpathlineto{\pgfqpoint{2.127897in}{0.315988in}}%
\pgfpathlineto{\pgfqpoint{2.127897in}{0.420167in}}%
\pgfpathlineto{\pgfqpoint{1.980983in}{0.420167in}}%
\pgfpathlineto{\pgfqpoint{1.980983in}{0.315988in}}%
\pgfpathclose%
\pgfusepath{fill}%
\end{pgfscope}%
\begin{pgfscope}%
\pgfpathrectangle{\pgfqpoint{0.394308in}{0.315988in}}{\pgfqpoint{2.585692in}{1.814012in}}%
\pgfusepath{clip}%
\pgfsetbuttcap%
\pgfsetmiterjoin%
\definecolor{currentfill}{rgb}{0.121569,0.466667,0.705882}%
\pgfsetfillcolor{currentfill}%
\pgfsetfillopacity{0.500000}%
\pgfsetlinewidth{0.000000pt}%
\definecolor{currentstroke}{rgb}{0.000000,0.000000,0.000000}%
\pgfsetstrokecolor{currentstroke}%
\pgfsetstrokeopacity{0.500000}%
\pgfsetdash{}{0pt}%
\pgfpathmoveto{\pgfqpoint{2.127897in}{0.315988in}}%
\pgfpathlineto{\pgfqpoint{2.274811in}{0.315988in}}%
\pgfpathlineto{\pgfqpoint{2.274811in}{0.368077in}}%
\pgfpathlineto{\pgfqpoint{2.127897in}{0.368077in}}%
\pgfpathlineto{\pgfqpoint{2.127897in}{0.315988in}}%
\pgfpathclose%
\pgfusepath{fill}%
\end{pgfscope}%
\begin{pgfscope}%
\pgfpathrectangle{\pgfqpoint{0.394308in}{0.315988in}}{\pgfqpoint{2.585692in}{1.814012in}}%
\pgfusepath{clip}%
\pgfsetbuttcap%
\pgfsetmiterjoin%
\definecolor{currentfill}{rgb}{0.121569,0.466667,0.705882}%
\pgfsetfillcolor{currentfill}%
\pgfsetfillopacity{0.500000}%
\pgfsetlinewidth{0.000000pt}%
\definecolor{currentstroke}{rgb}{0.000000,0.000000,0.000000}%
\pgfsetstrokecolor{currentstroke}%
\pgfsetstrokeopacity{0.500000}%
\pgfsetdash{}{0pt}%
\pgfpathmoveto{\pgfqpoint{2.274811in}{0.315988in}}%
\pgfpathlineto{\pgfqpoint{2.421726in}{0.315988in}}%
\pgfpathlineto{\pgfqpoint{2.421726in}{0.585116in}}%
\pgfpathlineto{\pgfqpoint{2.274811in}{0.585116in}}%
\pgfpathlineto{\pgfqpoint{2.274811in}{0.315988in}}%
\pgfpathclose%
\pgfusepath{fill}%
\end{pgfscope}%
\begin{pgfscope}%
\pgfpathrectangle{\pgfqpoint{0.394308in}{0.315988in}}{\pgfqpoint{2.585692in}{1.814012in}}%
\pgfusepath{clip}%
\pgfsetbuttcap%
\pgfsetmiterjoin%
\definecolor{currentfill}{rgb}{0.121569,0.466667,0.705882}%
\pgfsetfillcolor{currentfill}%
\pgfsetfillopacity{0.500000}%
\pgfsetlinewidth{0.000000pt}%
\definecolor{currentstroke}{rgb}{0.000000,0.000000,0.000000}%
\pgfsetstrokecolor{currentstroke}%
\pgfsetstrokeopacity{0.500000}%
\pgfsetdash{}{0pt}%
\pgfpathmoveto{\pgfqpoint{2.421726in}{0.315988in}}%
\pgfpathlineto{\pgfqpoint{2.568640in}{0.315988in}}%
\pgfpathlineto{\pgfqpoint{2.568640in}{0.515664in}}%
\pgfpathlineto{\pgfqpoint{2.421726in}{0.515664in}}%
\pgfpathlineto{\pgfqpoint{2.421726in}{0.315988in}}%
\pgfpathclose%
\pgfusepath{fill}%
\end{pgfscope}%
\begin{pgfscope}%
\pgfpathrectangle{\pgfqpoint{0.394308in}{0.315988in}}{\pgfqpoint{2.585692in}{1.814012in}}%
\pgfusepath{clip}%
\pgfsetbuttcap%
\pgfsetmiterjoin%
\definecolor{currentfill}{rgb}{0.121569,0.466667,0.705882}%
\pgfsetfillcolor{currentfill}%
\pgfsetfillopacity{0.500000}%
\pgfsetlinewidth{0.000000pt}%
\definecolor{currentstroke}{rgb}{0.000000,0.000000,0.000000}%
\pgfsetstrokecolor{currentstroke}%
\pgfsetstrokeopacity{0.500000}%
\pgfsetdash{}{0pt}%
\pgfpathmoveto{\pgfqpoint{2.568640in}{0.315988in}}%
\pgfpathlineto{\pgfqpoint{2.715554in}{0.315988in}}%
\pgfpathlineto{\pgfqpoint{2.715554in}{0.315988in}}%
\pgfpathlineto{\pgfqpoint{2.568640in}{0.315988in}}%
\pgfpathlineto{\pgfqpoint{2.568640in}{0.315988in}}%
\pgfpathclose%
\pgfusepath{fill}%
\end{pgfscope}%
\begin{pgfscope}%
\pgfpathrectangle{\pgfqpoint{0.394308in}{0.315988in}}{\pgfqpoint{2.585692in}{1.814012in}}%
\pgfusepath{clip}%
\pgfsetbuttcap%
\pgfsetmiterjoin%
\definecolor{currentfill}{rgb}{0.121569,0.466667,0.705882}%
\pgfsetfillcolor{currentfill}%
\pgfsetfillopacity{0.500000}%
\pgfsetlinewidth{0.000000pt}%
\definecolor{currentstroke}{rgb}{0.000000,0.000000,0.000000}%
\pgfsetstrokecolor{currentstroke}%
\pgfsetstrokeopacity{0.500000}%
\pgfsetdash{}{0pt}%
\pgfpathmoveto{\pgfqpoint{2.715554in}{0.315988in}}%
\pgfpathlineto{\pgfqpoint{2.862469in}{0.315988in}}%
\pgfpathlineto{\pgfqpoint{2.862469in}{2.043618in}}%
\pgfpathlineto{\pgfqpoint{2.715554in}{2.043618in}}%
\pgfpathlineto{\pgfqpoint{2.715554in}{0.315988in}}%
\pgfpathclose%
\pgfusepath{fill}%
\end{pgfscope}%
\begin{pgfscope}%
\pgfpathrectangle{\pgfqpoint{0.394308in}{0.315988in}}{\pgfqpoint{2.585692in}{1.814012in}}%
\pgfusepath{clip}%
\pgfsetbuttcap%
\pgfsetmiterjoin%
\definecolor{currentfill}{rgb}{1.000000,0.498039,0.054902}%
\pgfsetfillcolor{currentfill}%
\pgfsetfillopacity{0.500000}%
\pgfsetlinewidth{0.000000pt}%
\definecolor{currentstroke}{rgb}{0.000000,0.000000,0.000000}%
\pgfsetstrokecolor{currentstroke}%
\pgfsetstrokeopacity{0.500000}%
\pgfsetdash{}{0pt}%
\pgfpathmoveto{\pgfqpoint{0.511840in}{0.315988in}}%
\pgfpathlineto{\pgfqpoint{0.658754in}{0.315988in}}%
\pgfpathlineto{\pgfqpoint{0.658754in}{0.324669in}}%
\pgfpathlineto{\pgfqpoint{0.511840in}{0.324669in}}%
\pgfpathlineto{\pgfqpoint{0.511840in}{0.315988in}}%
\pgfpathclose%
\pgfusepath{fill}%
\end{pgfscope}%
\begin{pgfscope}%
\pgfpathrectangle{\pgfqpoint{0.394308in}{0.315988in}}{\pgfqpoint{2.585692in}{1.814012in}}%
\pgfusepath{clip}%
\pgfsetbuttcap%
\pgfsetmiterjoin%
\definecolor{currentfill}{rgb}{1.000000,0.498039,0.054902}%
\pgfsetfillcolor{currentfill}%
\pgfsetfillopacity{0.500000}%
\pgfsetlinewidth{0.000000pt}%
\definecolor{currentstroke}{rgb}{0.000000,0.000000,0.000000}%
\pgfsetstrokecolor{currentstroke}%
\pgfsetstrokeopacity{0.500000}%
\pgfsetdash{}{0pt}%
\pgfpathmoveto{\pgfqpoint{0.658754in}{0.315988in}}%
\pgfpathlineto{\pgfqpoint{0.805668in}{0.315988in}}%
\pgfpathlineto{\pgfqpoint{0.805668in}{0.315988in}}%
\pgfpathlineto{\pgfqpoint{0.658754in}{0.315988in}}%
\pgfpathlineto{\pgfqpoint{0.658754in}{0.315988in}}%
\pgfpathclose%
\pgfusepath{fill}%
\end{pgfscope}%
\begin{pgfscope}%
\pgfpathrectangle{\pgfqpoint{0.394308in}{0.315988in}}{\pgfqpoint{2.585692in}{1.814012in}}%
\pgfusepath{clip}%
\pgfsetbuttcap%
\pgfsetmiterjoin%
\definecolor{currentfill}{rgb}{1.000000,0.498039,0.054902}%
\pgfsetfillcolor{currentfill}%
\pgfsetfillopacity{0.500000}%
\pgfsetlinewidth{0.000000pt}%
\definecolor{currentstroke}{rgb}{0.000000,0.000000,0.000000}%
\pgfsetstrokecolor{currentstroke}%
\pgfsetstrokeopacity{0.500000}%
\pgfsetdash{}{0pt}%
\pgfpathmoveto{\pgfqpoint{0.805668in}{0.315988in}}%
\pgfpathlineto{\pgfqpoint{0.952582in}{0.315988in}}%
\pgfpathlineto{\pgfqpoint{0.952582in}{0.315988in}}%
\pgfpathlineto{\pgfqpoint{0.805668in}{0.315988in}}%
\pgfpathlineto{\pgfqpoint{0.805668in}{0.315988in}}%
\pgfpathclose%
\pgfusepath{fill}%
\end{pgfscope}%
\begin{pgfscope}%
\pgfpathrectangle{\pgfqpoint{0.394308in}{0.315988in}}{\pgfqpoint{2.585692in}{1.814012in}}%
\pgfusepath{clip}%
\pgfsetbuttcap%
\pgfsetmiterjoin%
\definecolor{currentfill}{rgb}{1.000000,0.498039,0.054902}%
\pgfsetfillcolor{currentfill}%
\pgfsetfillopacity{0.500000}%
\pgfsetlinewidth{0.000000pt}%
\definecolor{currentstroke}{rgb}{0.000000,0.000000,0.000000}%
\pgfsetstrokecolor{currentstroke}%
\pgfsetstrokeopacity{0.500000}%
\pgfsetdash{}{0pt}%
\pgfpathmoveto{\pgfqpoint{0.952582in}{0.315988in}}%
\pgfpathlineto{\pgfqpoint{1.099497in}{0.315988in}}%
\pgfpathlineto{\pgfqpoint{1.099497in}{0.342032in}}%
\pgfpathlineto{\pgfqpoint{0.952582in}{0.342032in}}%
\pgfpathlineto{\pgfqpoint{0.952582in}{0.315988in}}%
\pgfpathclose%
\pgfusepath{fill}%
\end{pgfscope}%
\begin{pgfscope}%
\pgfpathrectangle{\pgfqpoint{0.394308in}{0.315988in}}{\pgfqpoint{2.585692in}{1.814012in}}%
\pgfusepath{clip}%
\pgfsetbuttcap%
\pgfsetmiterjoin%
\definecolor{currentfill}{rgb}{1.000000,0.498039,0.054902}%
\pgfsetfillcolor{currentfill}%
\pgfsetfillopacity{0.500000}%
\pgfsetlinewidth{0.000000pt}%
\definecolor{currentstroke}{rgb}{0.000000,0.000000,0.000000}%
\pgfsetstrokecolor{currentstroke}%
\pgfsetstrokeopacity{0.500000}%
\pgfsetdash{}{0pt}%
\pgfpathmoveto{\pgfqpoint{1.099497in}{0.315988in}}%
\pgfpathlineto{\pgfqpoint{1.246411in}{0.315988in}}%
\pgfpathlineto{\pgfqpoint{1.246411in}{0.350714in}}%
\pgfpathlineto{\pgfqpoint{1.099497in}{0.350714in}}%
\pgfpathlineto{\pgfqpoint{1.099497in}{0.315988in}}%
\pgfpathclose%
\pgfusepath{fill}%
\end{pgfscope}%
\begin{pgfscope}%
\pgfpathrectangle{\pgfqpoint{0.394308in}{0.315988in}}{\pgfqpoint{2.585692in}{1.814012in}}%
\pgfusepath{clip}%
\pgfsetbuttcap%
\pgfsetmiterjoin%
\definecolor{currentfill}{rgb}{1.000000,0.498039,0.054902}%
\pgfsetfillcolor{currentfill}%
\pgfsetfillopacity{0.500000}%
\pgfsetlinewidth{0.000000pt}%
\definecolor{currentstroke}{rgb}{0.000000,0.000000,0.000000}%
\pgfsetstrokecolor{currentstroke}%
\pgfsetstrokeopacity{0.500000}%
\pgfsetdash{}{0pt}%
\pgfpathmoveto{\pgfqpoint{1.246411in}{0.315988in}}%
\pgfpathlineto{\pgfqpoint{1.393325in}{0.315988in}}%
\pgfpathlineto{\pgfqpoint{1.393325in}{0.333351in}}%
\pgfpathlineto{\pgfqpoint{1.246411in}{0.333351in}}%
\pgfpathlineto{\pgfqpoint{1.246411in}{0.315988in}}%
\pgfpathclose%
\pgfusepath{fill}%
\end{pgfscope}%
\begin{pgfscope}%
\pgfpathrectangle{\pgfqpoint{0.394308in}{0.315988in}}{\pgfqpoint{2.585692in}{1.814012in}}%
\pgfusepath{clip}%
\pgfsetbuttcap%
\pgfsetmiterjoin%
\definecolor{currentfill}{rgb}{1.000000,0.498039,0.054902}%
\pgfsetfillcolor{currentfill}%
\pgfsetfillopacity{0.500000}%
\pgfsetlinewidth{0.000000pt}%
\definecolor{currentstroke}{rgb}{0.000000,0.000000,0.000000}%
\pgfsetstrokecolor{currentstroke}%
\pgfsetstrokeopacity{0.500000}%
\pgfsetdash{}{0pt}%
\pgfpathmoveto{\pgfqpoint{1.393325in}{0.315988in}}%
\pgfpathlineto{\pgfqpoint{1.540240in}{0.315988in}}%
\pgfpathlineto{\pgfqpoint{1.540240in}{0.333351in}}%
\pgfpathlineto{\pgfqpoint{1.393325in}{0.333351in}}%
\pgfpathlineto{\pgfqpoint{1.393325in}{0.315988in}}%
\pgfpathclose%
\pgfusepath{fill}%
\end{pgfscope}%
\begin{pgfscope}%
\pgfpathrectangle{\pgfqpoint{0.394308in}{0.315988in}}{\pgfqpoint{2.585692in}{1.814012in}}%
\pgfusepath{clip}%
\pgfsetbuttcap%
\pgfsetmiterjoin%
\definecolor{currentfill}{rgb}{1.000000,0.498039,0.054902}%
\pgfsetfillcolor{currentfill}%
\pgfsetfillopacity{0.500000}%
\pgfsetlinewidth{0.000000pt}%
\definecolor{currentstroke}{rgb}{0.000000,0.000000,0.000000}%
\pgfsetstrokecolor{currentstroke}%
\pgfsetstrokeopacity{0.500000}%
\pgfsetdash{}{0pt}%
\pgfpathmoveto{\pgfqpoint{1.540240in}{0.315988in}}%
\pgfpathlineto{\pgfqpoint{1.687154in}{0.315988in}}%
\pgfpathlineto{\pgfqpoint{1.687154in}{0.446211in}}%
\pgfpathlineto{\pgfqpoint{1.540240in}{0.446211in}}%
\pgfpathlineto{\pgfqpoint{1.540240in}{0.315988in}}%
\pgfpathclose%
\pgfusepath{fill}%
\end{pgfscope}%
\begin{pgfscope}%
\pgfpathrectangle{\pgfqpoint{0.394308in}{0.315988in}}{\pgfqpoint{2.585692in}{1.814012in}}%
\pgfusepath{clip}%
\pgfsetbuttcap%
\pgfsetmiterjoin%
\definecolor{currentfill}{rgb}{1.000000,0.498039,0.054902}%
\pgfsetfillcolor{currentfill}%
\pgfsetfillopacity{0.500000}%
\pgfsetlinewidth{0.000000pt}%
\definecolor{currentstroke}{rgb}{0.000000,0.000000,0.000000}%
\pgfsetstrokecolor{currentstroke}%
\pgfsetstrokeopacity{0.500000}%
\pgfsetdash{}{0pt}%
\pgfpathmoveto{\pgfqpoint{1.687154in}{0.315988in}}%
\pgfpathlineto{\pgfqpoint{1.834068in}{0.315988in}}%
\pgfpathlineto{\pgfqpoint{1.834068in}{0.368077in}}%
\pgfpathlineto{\pgfqpoint{1.687154in}{0.368077in}}%
\pgfpathlineto{\pgfqpoint{1.687154in}{0.315988in}}%
\pgfpathclose%
\pgfusepath{fill}%
\end{pgfscope}%
\begin{pgfscope}%
\pgfpathrectangle{\pgfqpoint{0.394308in}{0.315988in}}{\pgfqpoint{2.585692in}{1.814012in}}%
\pgfusepath{clip}%
\pgfsetbuttcap%
\pgfsetmiterjoin%
\definecolor{currentfill}{rgb}{1.000000,0.498039,0.054902}%
\pgfsetfillcolor{currentfill}%
\pgfsetfillopacity{0.500000}%
\pgfsetlinewidth{0.000000pt}%
\definecolor{currentstroke}{rgb}{0.000000,0.000000,0.000000}%
\pgfsetstrokecolor{currentstroke}%
\pgfsetstrokeopacity{0.500000}%
\pgfsetdash{}{0pt}%
\pgfpathmoveto{\pgfqpoint{1.834068in}{0.315988in}}%
\pgfpathlineto{\pgfqpoint{1.980983in}{0.315988in}}%
\pgfpathlineto{\pgfqpoint{1.980983in}{0.385440in}}%
\pgfpathlineto{\pgfqpoint{1.834068in}{0.385440in}}%
\pgfpathlineto{\pgfqpoint{1.834068in}{0.315988in}}%
\pgfpathclose%
\pgfusepath{fill}%
\end{pgfscope}%
\begin{pgfscope}%
\pgfpathrectangle{\pgfqpoint{0.394308in}{0.315988in}}{\pgfqpoint{2.585692in}{1.814012in}}%
\pgfusepath{clip}%
\pgfsetbuttcap%
\pgfsetmiterjoin%
\definecolor{currentfill}{rgb}{1.000000,0.498039,0.054902}%
\pgfsetfillcolor{currentfill}%
\pgfsetfillopacity{0.500000}%
\pgfsetlinewidth{0.000000pt}%
\definecolor{currentstroke}{rgb}{0.000000,0.000000,0.000000}%
\pgfsetstrokecolor{currentstroke}%
\pgfsetstrokeopacity{0.500000}%
\pgfsetdash{}{0pt}%
\pgfpathmoveto{\pgfqpoint{1.980983in}{0.315988in}}%
\pgfpathlineto{\pgfqpoint{2.127897in}{0.315988in}}%
\pgfpathlineto{\pgfqpoint{2.127897in}{0.394122in}}%
\pgfpathlineto{\pgfqpoint{1.980983in}{0.394122in}}%
\pgfpathlineto{\pgfqpoint{1.980983in}{0.315988in}}%
\pgfpathclose%
\pgfusepath{fill}%
\end{pgfscope}%
\begin{pgfscope}%
\pgfpathrectangle{\pgfqpoint{0.394308in}{0.315988in}}{\pgfqpoint{2.585692in}{1.814012in}}%
\pgfusepath{clip}%
\pgfsetbuttcap%
\pgfsetmiterjoin%
\definecolor{currentfill}{rgb}{1.000000,0.498039,0.054902}%
\pgfsetfillcolor{currentfill}%
\pgfsetfillopacity{0.500000}%
\pgfsetlinewidth{0.000000pt}%
\definecolor{currentstroke}{rgb}{0.000000,0.000000,0.000000}%
\pgfsetstrokecolor{currentstroke}%
\pgfsetstrokeopacity{0.500000}%
\pgfsetdash{}{0pt}%
\pgfpathmoveto{\pgfqpoint{2.127897in}{0.315988in}}%
\pgfpathlineto{\pgfqpoint{2.274811in}{0.315988in}}%
\pgfpathlineto{\pgfqpoint{2.274811in}{0.515664in}}%
\pgfpathlineto{\pgfqpoint{2.127897in}{0.515664in}}%
\pgfpathlineto{\pgfqpoint{2.127897in}{0.315988in}}%
\pgfpathclose%
\pgfusepath{fill}%
\end{pgfscope}%
\begin{pgfscope}%
\pgfpathrectangle{\pgfqpoint{0.394308in}{0.315988in}}{\pgfqpoint{2.585692in}{1.814012in}}%
\pgfusepath{clip}%
\pgfsetbuttcap%
\pgfsetmiterjoin%
\definecolor{currentfill}{rgb}{1.000000,0.498039,0.054902}%
\pgfsetfillcolor{currentfill}%
\pgfsetfillopacity{0.500000}%
\pgfsetlinewidth{0.000000pt}%
\definecolor{currentstroke}{rgb}{0.000000,0.000000,0.000000}%
\pgfsetstrokecolor{currentstroke}%
\pgfsetstrokeopacity{0.500000}%
\pgfsetdash{}{0pt}%
\pgfpathmoveto{\pgfqpoint{2.274811in}{0.315988in}}%
\pgfpathlineto{\pgfqpoint{2.421726in}{0.315988in}}%
\pgfpathlineto{\pgfqpoint{2.421726in}{0.628524in}}%
\pgfpathlineto{\pgfqpoint{2.274811in}{0.628524in}}%
\pgfpathlineto{\pgfqpoint{2.274811in}{0.315988in}}%
\pgfpathclose%
\pgfusepath{fill}%
\end{pgfscope}%
\begin{pgfscope}%
\pgfpathrectangle{\pgfqpoint{0.394308in}{0.315988in}}{\pgfqpoint{2.585692in}{1.814012in}}%
\pgfusepath{clip}%
\pgfsetbuttcap%
\pgfsetmiterjoin%
\definecolor{currentfill}{rgb}{1.000000,0.498039,0.054902}%
\pgfsetfillcolor{currentfill}%
\pgfsetfillopacity{0.500000}%
\pgfsetlinewidth{0.000000pt}%
\definecolor{currentstroke}{rgb}{0.000000,0.000000,0.000000}%
\pgfsetstrokecolor{currentstroke}%
\pgfsetstrokeopacity{0.500000}%
\pgfsetdash{}{0pt}%
\pgfpathmoveto{\pgfqpoint{2.421726in}{0.315988in}}%
\pgfpathlineto{\pgfqpoint{2.568640in}{0.315988in}}%
\pgfpathlineto{\pgfqpoint{2.568640in}{0.472256in}}%
\pgfpathlineto{\pgfqpoint{2.421726in}{0.472256in}}%
\pgfpathlineto{\pgfqpoint{2.421726in}{0.315988in}}%
\pgfpathclose%
\pgfusepath{fill}%
\end{pgfscope}%
\begin{pgfscope}%
\pgfpathrectangle{\pgfqpoint{0.394308in}{0.315988in}}{\pgfqpoint{2.585692in}{1.814012in}}%
\pgfusepath{clip}%
\pgfsetbuttcap%
\pgfsetmiterjoin%
\definecolor{currentfill}{rgb}{1.000000,0.498039,0.054902}%
\pgfsetfillcolor{currentfill}%
\pgfsetfillopacity{0.500000}%
\pgfsetlinewidth{0.000000pt}%
\definecolor{currentstroke}{rgb}{0.000000,0.000000,0.000000}%
\pgfsetstrokecolor{currentstroke}%
\pgfsetstrokeopacity{0.500000}%
\pgfsetdash{}{0pt}%
\pgfpathmoveto{\pgfqpoint{2.568640in}{0.315988in}}%
\pgfpathlineto{\pgfqpoint{2.715554in}{0.315988in}}%
\pgfpathlineto{\pgfqpoint{2.715554in}{0.463574in}}%
\pgfpathlineto{\pgfqpoint{2.568640in}{0.463574in}}%
\pgfpathlineto{\pgfqpoint{2.568640in}{0.315988in}}%
\pgfpathclose%
\pgfusepath{fill}%
\end{pgfscope}%
\begin{pgfscope}%
\pgfpathrectangle{\pgfqpoint{0.394308in}{0.315988in}}{\pgfqpoint{2.585692in}{1.814012in}}%
\pgfusepath{clip}%
\pgfsetbuttcap%
\pgfsetmiterjoin%
\definecolor{currentfill}{rgb}{1.000000,0.498039,0.054902}%
\pgfsetfillcolor{currentfill}%
\pgfsetfillopacity{0.500000}%
\pgfsetlinewidth{0.000000pt}%
\definecolor{currentstroke}{rgb}{0.000000,0.000000,0.000000}%
\pgfsetstrokecolor{currentstroke}%
\pgfsetstrokeopacity{0.500000}%
\pgfsetdash{}{0pt}%
\pgfpathmoveto{\pgfqpoint{2.715554in}{0.315988in}}%
\pgfpathlineto{\pgfqpoint{2.862469in}{0.315988in}}%
\pgfpathlineto{\pgfqpoint{2.862469in}{1.626904in}}%
\pgfpathlineto{\pgfqpoint{2.715554in}{1.626904in}}%
\pgfpathlineto{\pgfqpoint{2.715554in}{0.315988in}}%
\pgfpathclose%
\pgfusepath{fill}%
\end{pgfscope}%
\begin{pgfscope}%
\pgfsetbuttcap%
\pgfsetroundjoin%
\definecolor{currentfill}{rgb}{0.000000,0.000000,0.000000}%
\pgfsetfillcolor{currentfill}%
\pgfsetlinewidth{0.803000pt}%
\definecolor{currentstroke}{rgb}{0.000000,0.000000,0.000000}%
\pgfsetstrokecolor{currentstroke}%
\pgfsetdash{}{0pt}%
\pgfsys@defobject{currentmarker}{\pgfqpoint{0.000000in}{-0.048611in}}{\pgfqpoint{0.000000in}{0.000000in}}{%
\pgfpathmoveto{\pgfqpoint{0.000000in}{0.000000in}}%
\pgfpathlineto{\pgfqpoint{0.000000in}{-0.048611in}}%
\pgfusepath{stroke,fill}%
}%
\begin{pgfscope}%
\pgfsys@transformshift{0.511840in}{0.315988in}%
\pgfsys@useobject{currentmarker}{}%
\end{pgfscope}%
\end{pgfscope}%
\begin{pgfscope}%
\definecolor{textcolor}{rgb}{0.000000,0.000000,0.000000}%
\pgfsetstrokecolor{textcolor}%
\pgfsetfillcolor{textcolor}%
\pgftext[x=0.511840in,y=0.218766in,,top]{\color{textcolor}\rmfamily\fontsize{8.000000}{9.600000}\selectfont \(\displaystyle {\ensuremath{-}3}\)}%
\end{pgfscope}%
\begin{pgfscope}%
\pgfsetbuttcap%
\pgfsetroundjoin%
\definecolor{currentfill}{rgb}{0.000000,0.000000,0.000000}%
\pgfsetfillcolor{currentfill}%
\pgfsetlinewidth{0.803000pt}%
\definecolor{currentstroke}{rgb}{0.000000,0.000000,0.000000}%
\pgfsetstrokecolor{currentstroke}%
\pgfsetdash{}{0pt}%
\pgfsys@defobject{currentmarker}{\pgfqpoint{0.000000in}{-0.048611in}}{\pgfqpoint{0.000000in}{0.000000in}}{%
\pgfpathmoveto{\pgfqpoint{0.000000in}{0.000000in}}%
\pgfpathlineto{\pgfqpoint{0.000000in}{-0.048611in}}%
\pgfusepath{stroke,fill}%
}%
\begin{pgfscope}%
\pgfsys@transformshift{1.099497in}{0.315988in}%
\pgfsys@useobject{currentmarker}{}%
\end{pgfscope}%
\end{pgfscope}%
\begin{pgfscope}%
\definecolor{textcolor}{rgb}{0.000000,0.000000,0.000000}%
\pgfsetstrokecolor{textcolor}%
\pgfsetfillcolor{textcolor}%
\pgftext[x=1.099497in,y=0.218766in,,top]{\color{textcolor}\rmfamily\fontsize{8.000000}{9.600000}\selectfont \(\displaystyle {\ensuremath{-}2}\)}%
\end{pgfscope}%
\begin{pgfscope}%
\pgfsetbuttcap%
\pgfsetroundjoin%
\definecolor{currentfill}{rgb}{0.000000,0.000000,0.000000}%
\pgfsetfillcolor{currentfill}%
\pgfsetlinewidth{0.803000pt}%
\definecolor{currentstroke}{rgb}{0.000000,0.000000,0.000000}%
\pgfsetstrokecolor{currentstroke}%
\pgfsetdash{}{0pt}%
\pgfsys@defobject{currentmarker}{\pgfqpoint{0.000000in}{-0.048611in}}{\pgfqpoint{0.000000in}{0.000000in}}{%
\pgfpathmoveto{\pgfqpoint{0.000000in}{0.000000in}}%
\pgfpathlineto{\pgfqpoint{0.000000in}{-0.048611in}}%
\pgfusepath{stroke,fill}%
}%
\begin{pgfscope}%
\pgfsys@transformshift{1.687154in}{0.315988in}%
\pgfsys@useobject{currentmarker}{}%
\end{pgfscope}%
\end{pgfscope}%
\begin{pgfscope}%
\definecolor{textcolor}{rgb}{0.000000,0.000000,0.000000}%
\pgfsetstrokecolor{textcolor}%
\pgfsetfillcolor{textcolor}%
\pgftext[x=1.687154in,y=0.218766in,,top]{\color{textcolor}\rmfamily\fontsize{8.000000}{9.600000}\selectfont \(\displaystyle {\ensuremath{-}1}\)}%
\end{pgfscope}%
\begin{pgfscope}%
\pgfsetbuttcap%
\pgfsetroundjoin%
\definecolor{currentfill}{rgb}{0.000000,0.000000,0.000000}%
\pgfsetfillcolor{currentfill}%
\pgfsetlinewidth{0.803000pt}%
\definecolor{currentstroke}{rgb}{0.000000,0.000000,0.000000}%
\pgfsetstrokecolor{currentstroke}%
\pgfsetdash{}{0pt}%
\pgfsys@defobject{currentmarker}{\pgfqpoint{0.000000in}{-0.048611in}}{\pgfqpoint{0.000000in}{0.000000in}}{%
\pgfpathmoveto{\pgfqpoint{0.000000in}{0.000000in}}%
\pgfpathlineto{\pgfqpoint{0.000000in}{-0.048611in}}%
\pgfusepath{stroke,fill}%
}%
\begin{pgfscope}%
\pgfsys@transformshift{2.274811in}{0.315988in}%
\pgfsys@useobject{currentmarker}{}%
\end{pgfscope}%
\end{pgfscope}%
\begin{pgfscope}%
\definecolor{textcolor}{rgb}{0.000000,0.000000,0.000000}%
\pgfsetstrokecolor{textcolor}%
\pgfsetfillcolor{textcolor}%
\pgftext[x=2.274811in,y=0.218766in,,top]{\color{textcolor}\rmfamily\fontsize{8.000000}{9.600000}\selectfont \(\displaystyle {0}\)}%
\end{pgfscope}%
\begin{pgfscope}%
\pgfsetbuttcap%
\pgfsetroundjoin%
\definecolor{currentfill}{rgb}{0.000000,0.000000,0.000000}%
\pgfsetfillcolor{currentfill}%
\pgfsetlinewidth{0.803000pt}%
\definecolor{currentstroke}{rgb}{0.000000,0.000000,0.000000}%
\pgfsetstrokecolor{currentstroke}%
\pgfsetdash{}{0pt}%
\pgfsys@defobject{currentmarker}{\pgfqpoint{0.000000in}{-0.048611in}}{\pgfqpoint{0.000000in}{0.000000in}}{%
\pgfpathmoveto{\pgfqpoint{0.000000in}{0.000000in}}%
\pgfpathlineto{\pgfqpoint{0.000000in}{-0.048611in}}%
\pgfusepath{stroke,fill}%
}%
\begin{pgfscope}%
\pgfsys@transformshift{2.862469in}{0.315988in}%
\pgfsys@useobject{currentmarker}{}%
\end{pgfscope}%
\end{pgfscope}%
\begin{pgfscope}%
\definecolor{textcolor}{rgb}{0.000000,0.000000,0.000000}%
\pgfsetstrokecolor{textcolor}%
\pgfsetfillcolor{textcolor}%
\pgftext[x=2.862469in,y=0.218766in,,top]{\color{textcolor}\rmfamily\fontsize{8.000000}{9.600000}\selectfont \(\displaystyle {1}\)}%
\end{pgfscope}%
\begin{pgfscope}%
\pgfsetbuttcap%
\pgfsetroundjoin%
\definecolor{currentfill}{rgb}{0.000000,0.000000,0.000000}%
\pgfsetfillcolor{currentfill}%
\pgfsetlinewidth{0.803000pt}%
\definecolor{currentstroke}{rgb}{0.000000,0.000000,0.000000}%
\pgfsetstrokecolor{currentstroke}%
\pgfsetdash{}{0pt}%
\pgfsys@defobject{currentmarker}{\pgfqpoint{-0.048611in}{0.000000in}}{\pgfqpoint{-0.000000in}{0.000000in}}{%
\pgfpathmoveto{\pgfqpoint{-0.000000in}{0.000000in}}%
\pgfpathlineto{\pgfqpoint{-0.048611in}{0.000000in}}%
\pgfusepath{stroke,fill}%
}%
\begin{pgfscope}%
\pgfsys@transformshift{0.394308in}{0.315988in}%
\pgfsys@useobject{currentmarker}{}%
\end{pgfscope}%
\end{pgfscope}%
\begin{pgfscope}%
\definecolor{textcolor}{rgb}{0.000000,0.000000,0.000000}%
\pgfsetstrokecolor{textcolor}%
\pgfsetfillcolor{textcolor}%
\pgftext[x=0.238057in, y=0.277408in, left, base]{\color{textcolor}\rmfamily\fontsize{8.000000}{9.600000}\selectfont \(\displaystyle {0}\)}%
\end{pgfscope}%
\begin{pgfscope}%
\pgfsetbuttcap%
\pgfsetroundjoin%
\definecolor{currentfill}{rgb}{0.000000,0.000000,0.000000}%
\pgfsetfillcolor{currentfill}%
\pgfsetlinewidth{0.803000pt}%
\definecolor{currentstroke}{rgb}{0.000000,0.000000,0.000000}%
\pgfsetstrokecolor{currentstroke}%
\pgfsetdash{}{0pt}%
\pgfsys@defobject{currentmarker}{\pgfqpoint{-0.048611in}{0.000000in}}{\pgfqpoint{-0.000000in}{0.000000in}}{%
\pgfpathmoveto{\pgfqpoint{-0.000000in}{0.000000in}}%
\pgfpathlineto{\pgfqpoint{-0.048611in}{0.000000in}}%
\pgfusepath{stroke,fill}%
}%
\begin{pgfscope}%
\pgfsys@transformshift{0.394308in}{0.750066in}%
\pgfsys@useobject{currentmarker}{}%
\end{pgfscope}%
\end{pgfscope}%
\begin{pgfscope}%
\definecolor{textcolor}{rgb}{0.000000,0.000000,0.000000}%
\pgfsetstrokecolor{textcolor}%
\pgfsetfillcolor{textcolor}%
\pgftext[x=0.179029in, y=0.711486in, left, base]{\color{textcolor}\rmfamily\fontsize{8.000000}{9.600000}\selectfont \(\displaystyle {50}\)}%
\end{pgfscope}%
\begin{pgfscope}%
\pgfsetbuttcap%
\pgfsetroundjoin%
\definecolor{currentfill}{rgb}{0.000000,0.000000,0.000000}%
\pgfsetfillcolor{currentfill}%
\pgfsetlinewidth{0.803000pt}%
\definecolor{currentstroke}{rgb}{0.000000,0.000000,0.000000}%
\pgfsetstrokecolor{currentstroke}%
\pgfsetdash{}{0pt}%
\pgfsys@defobject{currentmarker}{\pgfqpoint{-0.048611in}{0.000000in}}{\pgfqpoint{-0.000000in}{0.000000in}}{%
\pgfpathmoveto{\pgfqpoint{-0.000000in}{0.000000in}}%
\pgfpathlineto{\pgfqpoint{-0.048611in}{0.000000in}}%
\pgfusepath{stroke,fill}%
}%
\begin{pgfscope}%
\pgfsys@transformshift{0.394308in}{1.184144in}%
\pgfsys@useobject{currentmarker}{}%
\end{pgfscope}%
\end{pgfscope}%
\begin{pgfscope}%
\definecolor{textcolor}{rgb}{0.000000,0.000000,0.000000}%
\pgfsetstrokecolor{textcolor}%
\pgfsetfillcolor{textcolor}%
\pgftext[x=0.120000in, y=1.145564in, left, base]{\color{textcolor}\rmfamily\fontsize{8.000000}{9.600000}\selectfont \(\displaystyle {100}\)}%
\end{pgfscope}%
\begin{pgfscope}%
\pgfsetbuttcap%
\pgfsetroundjoin%
\definecolor{currentfill}{rgb}{0.000000,0.000000,0.000000}%
\pgfsetfillcolor{currentfill}%
\pgfsetlinewidth{0.803000pt}%
\definecolor{currentstroke}{rgb}{0.000000,0.000000,0.000000}%
\pgfsetstrokecolor{currentstroke}%
\pgfsetdash{}{0pt}%
\pgfsys@defobject{currentmarker}{\pgfqpoint{-0.048611in}{0.000000in}}{\pgfqpoint{-0.000000in}{0.000000in}}{%
\pgfpathmoveto{\pgfqpoint{-0.000000in}{0.000000in}}%
\pgfpathlineto{\pgfqpoint{-0.048611in}{0.000000in}}%
\pgfusepath{stroke,fill}%
}%
\begin{pgfscope}%
\pgfsys@transformshift{0.394308in}{1.618222in}%
\pgfsys@useobject{currentmarker}{}%
\end{pgfscope}%
\end{pgfscope}%
\begin{pgfscope}%
\definecolor{textcolor}{rgb}{0.000000,0.000000,0.000000}%
\pgfsetstrokecolor{textcolor}%
\pgfsetfillcolor{textcolor}%
\pgftext[x=0.120000in, y=1.579642in, left, base]{\color{textcolor}\rmfamily\fontsize{8.000000}{9.600000}\selectfont \(\displaystyle {150}\)}%
\end{pgfscope}%
\begin{pgfscope}%
\pgfsetbuttcap%
\pgfsetroundjoin%
\definecolor{currentfill}{rgb}{0.000000,0.000000,0.000000}%
\pgfsetfillcolor{currentfill}%
\pgfsetlinewidth{0.803000pt}%
\definecolor{currentstroke}{rgb}{0.000000,0.000000,0.000000}%
\pgfsetstrokecolor{currentstroke}%
\pgfsetdash{}{0pt}%
\pgfsys@defobject{currentmarker}{\pgfqpoint{-0.048611in}{0.000000in}}{\pgfqpoint{-0.000000in}{0.000000in}}{%
\pgfpathmoveto{\pgfqpoint{-0.000000in}{0.000000in}}%
\pgfpathlineto{\pgfqpoint{-0.048611in}{0.000000in}}%
\pgfusepath{stroke,fill}%
}%
\begin{pgfscope}%
\pgfsys@transformshift{0.394308in}{2.052300in}%
\pgfsys@useobject{currentmarker}{}%
\end{pgfscope}%
\end{pgfscope}%
\begin{pgfscope}%
\definecolor{textcolor}{rgb}{0.000000,0.000000,0.000000}%
\pgfsetstrokecolor{textcolor}%
\pgfsetfillcolor{textcolor}%
\pgftext[x=0.120000in, y=2.013720in, left, base]{\color{textcolor}\rmfamily\fontsize{8.000000}{9.600000}\selectfont \(\displaystyle {200}\)}%
\end{pgfscope}%
\begin{pgfscope}%
\pgfsetrectcap%
\pgfsetmiterjoin%
\pgfsetlinewidth{0.803000pt}%
\definecolor{currentstroke}{rgb}{0.000000,0.000000,0.000000}%
\pgfsetstrokecolor{currentstroke}%
\pgfsetdash{}{0pt}%
\pgfpathmoveto{\pgfqpoint{0.394308in}{0.315988in}}%
\pgfpathlineto{\pgfqpoint{0.394308in}{2.130000in}}%
\pgfusepath{stroke}%
\end{pgfscope}%
\begin{pgfscope}%
\pgfsetrectcap%
\pgfsetmiterjoin%
\pgfsetlinewidth{0.803000pt}%
\definecolor{currentstroke}{rgb}{0.000000,0.000000,0.000000}%
\pgfsetstrokecolor{currentstroke}%
\pgfsetdash{}{0pt}%
\pgfpathmoveto{\pgfqpoint{2.980000in}{0.315988in}}%
\pgfpathlineto{\pgfqpoint{2.980000in}{2.130000in}}%
\pgfusepath{stroke}%
\end{pgfscope}%
\begin{pgfscope}%
\pgfsetrectcap%
\pgfsetmiterjoin%
\pgfsetlinewidth{0.803000pt}%
\definecolor{currentstroke}{rgb}{0.000000,0.000000,0.000000}%
\pgfsetstrokecolor{currentstroke}%
\pgfsetdash{}{0pt}%
\pgfpathmoveto{\pgfqpoint{0.394308in}{0.315988in}}%
\pgfpathlineto{\pgfqpoint{2.980000in}{0.315988in}}%
\pgfusepath{stroke}%
\end{pgfscope}%
\begin{pgfscope}%
\pgfsetrectcap%
\pgfsetmiterjoin%
\pgfsetlinewidth{0.803000pt}%
\definecolor{currentstroke}{rgb}{0.000000,0.000000,0.000000}%
\pgfsetstrokecolor{currentstroke}%
\pgfsetdash{}{0pt}%
\pgfpathmoveto{\pgfqpoint{0.394308in}{2.130000in}}%
\pgfpathlineto{\pgfqpoint{2.980000in}{2.130000in}}%
\pgfusepath{stroke}%
\end{pgfscope}%
\begin{pgfscope}%
\pgfsetbuttcap%
\pgfsetmiterjoin%
\definecolor{currentfill}{rgb}{1.000000,1.000000,1.000000}%
\pgfsetfillcolor{currentfill}%
\pgfsetfillopacity{0.800000}%
\pgfsetlinewidth{1.003750pt}%
\definecolor{currentstroke}{rgb}{0.800000,0.800000,0.800000}%
\pgfsetstrokecolor{currentstroke}%
\pgfsetstrokeopacity{0.800000}%
\pgfsetdash{}{0pt}%
\pgfpathmoveto{\pgfqpoint{0.472086in}{1.719506in}}%
\pgfpathlineto{\pgfqpoint{1.870075in}{1.719506in}}%
\pgfpathquadraticcurveto{\pgfqpoint{1.892298in}{1.719506in}}{\pgfqpoint{1.892298in}{1.741728in}}%
\pgfpathlineto{\pgfqpoint{1.892298in}{2.052222in}}%
\pgfpathquadraticcurveto{\pgfqpoint{1.892298in}{2.074444in}}{\pgfqpoint{1.870075in}{2.074444in}}%
\pgfpathlineto{\pgfqpoint{0.472086in}{2.074444in}}%
\pgfpathquadraticcurveto{\pgfqpoint{0.449864in}{2.074444in}}{\pgfqpoint{0.449864in}{2.052222in}}%
\pgfpathlineto{\pgfqpoint{0.449864in}{1.741728in}}%
\pgfpathquadraticcurveto{\pgfqpoint{0.449864in}{1.719506in}}{\pgfqpoint{0.472086in}{1.719506in}}%
\pgfpathlineto{\pgfqpoint{0.472086in}{1.719506in}}%
\pgfpathclose%
\pgfusepath{stroke,fill}%
\end{pgfscope}%
\begin{pgfscope}%
\pgfsetbuttcap%
\pgfsetmiterjoin%
\definecolor{currentfill}{rgb}{0.121569,0.466667,0.705882}%
\pgfsetfillcolor{currentfill}%
\pgfsetfillopacity{0.500000}%
\pgfsetlinewidth{0.000000pt}%
\definecolor{currentstroke}{rgb}{0.000000,0.000000,0.000000}%
\pgfsetstrokecolor{currentstroke}%
\pgfsetstrokeopacity{0.500000}%
\pgfsetdash{}{0pt}%
\pgfpathmoveto{\pgfqpoint{0.494308in}{1.952222in}}%
\pgfpathlineto{\pgfqpoint{0.716530in}{1.952222in}}%
\pgfpathlineto{\pgfqpoint{0.716530in}{2.030000in}}%
\pgfpathlineto{\pgfqpoint{0.494308in}{2.030000in}}%
\pgfpathlineto{\pgfqpoint{0.494308in}{1.952222in}}%
\pgfpathclose%
\pgfusepath{fill}%
\end{pgfscope}%
\begin{pgfscope}%
\definecolor{textcolor}{rgb}{0.000000,0.000000,0.000000}%
\pgfsetstrokecolor{textcolor}%
\pgfsetfillcolor{textcolor}%
\pgftext[x=0.805419in,y=1.952222in,left,base]{\color{textcolor}\rmfamily\fontsize{8.000000}{9.600000}\selectfont Follows Instruction?}%
\end{pgfscope}%
\begin{pgfscope}%
\pgfsetbuttcap%
\pgfsetmiterjoin%
\definecolor{currentfill}{rgb}{1.000000,0.498039,0.054902}%
\pgfsetfillcolor{currentfill}%
\pgfsetfillopacity{0.500000}%
\pgfsetlinewidth{0.000000pt}%
\definecolor{currentstroke}{rgb}{0.000000,0.000000,0.000000}%
\pgfsetstrokecolor{currentstroke}%
\pgfsetstrokeopacity{0.500000}%
\pgfsetdash{}{0pt}%
\pgfpathmoveto{\pgfqpoint{0.494308in}{1.791728in}}%
\pgfpathlineto{\pgfqpoint{0.716530in}{1.791728in}}%
\pgfpathlineto{\pgfqpoint{0.716530in}{1.869506in}}%
\pgfpathlineto{\pgfqpoint{0.494308in}{1.869506in}}%
\pgfpathlineto{\pgfqpoint{0.494308in}{1.791728in}}%
\pgfpathclose%
\pgfusepath{fill}%
\end{pgfscope}%
\begin{pgfscope}%
\definecolor{textcolor}{rgb}{0.000000,0.000000,0.000000}%
\pgfsetstrokecolor{textcolor}%
\pgfsetfillcolor{textcolor}%
\pgftext[x=0.805419in,y=1.791728in,left,base]{\color{textcolor}\rmfamily\fontsize{8.000000}{9.600000}\selectfont How Well? (1-5)}%
\end{pgfscope}%
\end{pgfpicture}%
\makeatother%
\endgroup%

%% file: rater_question_distrib_histogram.pgf
%% Creator: Matplotlib, PGF backend
%%
%% To include the figure in your LaTeX document, write
%%   \input{<filename>.pgf}
%%
%% Make sure the required packages are loaded in your preamble
%%   \usepackage{pgf}
%%
%% Also ensure that all the required font packages are loaded; for instance,
%% the lmodern package is sometimes necessary when using math font.
%%   \usepackage{lmodern}
%%
%% Figures using additional raster images can only be included by \input if
%% they are in the same directory as the main LaTeX file. For loading figures
%% from other directories you can use the `import` package
%%   \usepackage{import}
%%
%% and then include the figures with
%%   \import{<path to file>}{<filename>.pgf}
%%
%% Matplotlib used the following preamble
%%   
%%   \makeatletter\@ifpackageloaded{underscore}{}{\usepackage[strings]{underscore}}\makeatother
%%
\begingroup%
\makeatletter%
\begin{pgfpicture}%
\pgfpathrectangle{\pgfpointorigin}{\pgfqpoint{3.250000in}{2.500000in}}%
\pgfusepath{use as bounding box, clip}%
\begin{pgfscope}%
\pgfsetbuttcap%
\pgfsetmiterjoin%
\pgfsetlinewidth{0.000000pt}%
\definecolor{currentstroke}{rgb}{1.000000,1.000000,1.000000}%
\pgfsetstrokecolor{currentstroke}%
\pgfsetstrokeopacity{0.000000}%
\pgfsetdash{}{0pt}%
\pgfpathmoveto{\pgfqpoint{0.000000in}{0.000000in}}%
\pgfpathlineto{\pgfqpoint{3.250000in}{0.000000in}}%
\pgfpathlineto{\pgfqpoint{3.250000in}{2.500000in}}%
\pgfpathlineto{\pgfqpoint{0.000000in}{2.500000in}}%
\pgfpathlineto{\pgfqpoint{0.000000in}{0.000000in}}%
\pgfpathclose%
\pgfusepath{}%
\end{pgfscope}%
\begin{pgfscope}%
\pgfsetbuttcap%
\pgfsetmiterjoin%
\definecolor{currentfill}{rgb}{1.000000,1.000000,1.000000}%
\pgfsetfillcolor{currentfill}%
\pgfsetlinewidth{0.000000pt}%
\definecolor{currentstroke}{rgb}{0.000000,0.000000,0.000000}%
\pgfsetstrokecolor{currentstroke}%
\pgfsetstrokeopacity{0.000000}%
\pgfsetdash{}{0pt}%
\pgfpathmoveto{\pgfqpoint{0.394308in}{0.315988in}}%
\pgfpathlineto{\pgfqpoint{3.130000in}{0.315988in}}%
\pgfpathlineto{\pgfqpoint{3.130000in}{2.371115in}}%
\pgfpathlineto{\pgfqpoint{0.394308in}{2.371115in}}%
\pgfpathlineto{\pgfqpoint{0.394308in}{0.315988in}}%
\pgfpathclose%
\pgfusepath{fill}%
\end{pgfscope}%
\begin{pgfscope}%
\pgfpathrectangle{\pgfqpoint{0.394308in}{0.315988in}}{\pgfqpoint{2.735692in}{2.055128in}}%
\pgfusepath{clip}%
\pgfsetbuttcap%
\pgfsetmiterjoin%
\definecolor{currentfill}{rgb}{0.121569,0.466667,0.705882}%
\pgfsetfillcolor{currentfill}%
\pgfsetlinewidth{0.000000pt}%
\definecolor{currentstroke}{rgb}{0.000000,0.000000,0.000000}%
\pgfsetstrokecolor{currentstroke}%
\pgfsetstrokeopacity{0.000000}%
\pgfsetdash{}{0pt}%
\pgfpathmoveto{\pgfqpoint{0.518658in}{0.315988in}}%
\pgfpathlineto{\pgfqpoint{0.662831in}{0.315988in}}%
\pgfpathlineto{\pgfqpoint{0.662831in}{2.273252in}}%
\pgfpathlineto{\pgfqpoint{0.518658in}{2.273252in}}%
\pgfpathlineto{\pgfqpoint{0.518658in}{0.315988in}}%
\pgfpathclose%
\pgfusepath{fill}%
\end{pgfscope}%
\begin{pgfscope}%
\pgfpathrectangle{\pgfqpoint{0.394308in}{0.315988in}}{\pgfqpoint{2.735692in}{2.055128in}}%
\pgfusepath{clip}%
\pgfsetbuttcap%
\pgfsetmiterjoin%
\definecolor{currentfill}{rgb}{0.121569,0.466667,0.705882}%
\pgfsetfillcolor{currentfill}%
\pgfsetlinewidth{0.000000pt}%
\definecolor{currentstroke}{rgb}{0.000000,0.000000,0.000000}%
\pgfsetstrokecolor{currentstroke}%
\pgfsetstrokeopacity{0.000000}%
\pgfsetdash{}{0pt}%
\pgfpathmoveto{\pgfqpoint{0.698875in}{0.315988in}}%
\pgfpathlineto{\pgfqpoint{0.843048in}{0.315988in}}%
\pgfpathlineto{\pgfqpoint{0.843048in}{1.995929in}}%
\pgfpathlineto{\pgfqpoint{0.698875in}{1.995929in}}%
\pgfpathlineto{\pgfqpoint{0.698875in}{0.315988in}}%
\pgfpathclose%
\pgfusepath{fill}%
\end{pgfscope}%
\begin{pgfscope}%
\pgfpathrectangle{\pgfqpoint{0.394308in}{0.315988in}}{\pgfqpoint{2.735692in}{2.055128in}}%
\pgfusepath{clip}%
\pgfsetbuttcap%
\pgfsetmiterjoin%
\definecolor{currentfill}{rgb}{0.121569,0.466667,0.705882}%
\pgfsetfillcolor{currentfill}%
\pgfsetlinewidth{0.000000pt}%
\definecolor{currentstroke}{rgb}{0.000000,0.000000,0.000000}%
\pgfsetstrokecolor{currentstroke}%
\pgfsetstrokeopacity{0.000000}%
\pgfsetdash{}{0pt}%
\pgfpathmoveto{\pgfqpoint{0.879091in}{0.315988in}}%
\pgfpathlineto{\pgfqpoint{1.023265in}{0.315988in}}%
\pgfpathlineto{\pgfqpoint{1.023265in}{1.979929in}}%
\pgfpathlineto{\pgfqpoint{0.879091in}{1.979929in}}%
\pgfpathlineto{\pgfqpoint{0.879091in}{0.315988in}}%
\pgfpathclose%
\pgfusepath{fill}%
\end{pgfscope}%
\begin{pgfscope}%
\pgfpathrectangle{\pgfqpoint{0.394308in}{0.315988in}}{\pgfqpoint{2.735692in}{2.055128in}}%
\pgfusepath{clip}%
\pgfsetbuttcap%
\pgfsetmiterjoin%
\definecolor{currentfill}{rgb}{0.121569,0.466667,0.705882}%
\pgfsetfillcolor{currentfill}%
\pgfsetlinewidth{0.000000pt}%
\definecolor{currentstroke}{rgb}{0.000000,0.000000,0.000000}%
\pgfsetstrokecolor{currentstroke}%
\pgfsetstrokeopacity{0.000000}%
\pgfsetdash{}{0pt}%
\pgfpathmoveto{\pgfqpoint{1.059308in}{0.315988in}}%
\pgfpathlineto{\pgfqpoint{1.203482in}{0.315988in}}%
\pgfpathlineto{\pgfqpoint{1.203482in}{1.953263in}}%
\pgfpathlineto{\pgfqpoint{1.059308in}{1.953263in}}%
\pgfpathlineto{\pgfqpoint{1.059308in}{0.315988in}}%
\pgfpathclose%
\pgfusepath{fill}%
\end{pgfscope}%
\begin{pgfscope}%
\pgfpathrectangle{\pgfqpoint{0.394308in}{0.315988in}}{\pgfqpoint{2.735692in}{2.055128in}}%
\pgfusepath{clip}%
\pgfsetbuttcap%
\pgfsetmiterjoin%
\definecolor{currentfill}{rgb}{0.121569,0.466667,0.705882}%
\pgfsetfillcolor{currentfill}%
\pgfsetlinewidth{0.000000pt}%
\definecolor{currentstroke}{rgb}{0.000000,0.000000,0.000000}%
\pgfsetstrokecolor{currentstroke}%
\pgfsetstrokeopacity{0.000000}%
\pgfsetdash{}{0pt}%
\pgfpathmoveto{\pgfqpoint{1.239525in}{0.315988in}}%
\pgfpathlineto{\pgfqpoint{1.383699in}{0.315988in}}%
\pgfpathlineto{\pgfqpoint{1.383699in}{1.825268in}}%
\pgfpathlineto{\pgfqpoint{1.239525in}{1.825268in}}%
\pgfpathlineto{\pgfqpoint{1.239525in}{0.315988in}}%
\pgfpathclose%
\pgfusepath{fill}%
\end{pgfscope}%
\begin{pgfscope}%
\pgfpathrectangle{\pgfqpoint{0.394308in}{0.315988in}}{\pgfqpoint{2.735692in}{2.055128in}}%
\pgfusepath{clip}%
\pgfsetbuttcap%
\pgfsetmiterjoin%
\definecolor{currentfill}{rgb}{0.121569,0.466667,0.705882}%
\pgfsetfillcolor{currentfill}%
\pgfsetlinewidth{0.000000pt}%
\definecolor{currentstroke}{rgb}{0.000000,0.000000,0.000000}%
\pgfsetstrokecolor{currentstroke}%
\pgfsetstrokeopacity{0.000000}%
\pgfsetdash{}{0pt}%
\pgfpathmoveto{\pgfqpoint{1.419742in}{0.315988in}}%
\pgfpathlineto{\pgfqpoint{1.563915in}{0.315988in}}%
\pgfpathlineto{\pgfqpoint{1.563915in}{1.579943in}}%
\pgfpathlineto{\pgfqpoint{1.419742in}{1.579943in}}%
\pgfpathlineto{\pgfqpoint{1.419742in}{0.315988in}}%
\pgfpathclose%
\pgfusepath{fill}%
\end{pgfscope}%
\begin{pgfscope}%
\pgfpathrectangle{\pgfqpoint{0.394308in}{0.315988in}}{\pgfqpoint{2.735692in}{2.055128in}}%
\pgfusepath{clip}%
\pgfsetbuttcap%
\pgfsetmiterjoin%
\definecolor{currentfill}{rgb}{0.121569,0.466667,0.705882}%
\pgfsetfillcolor{currentfill}%
\pgfsetlinewidth{0.000000pt}%
\definecolor{currentstroke}{rgb}{0.000000,0.000000,0.000000}%
\pgfsetstrokecolor{currentstroke}%
\pgfsetstrokeopacity{0.000000}%
\pgfsetdash{}{0pt}%
\pgfpathmoveto{\pgfqpoint{1.599959in}{0.315988in}}%
\pgfpathlineto{\pgfqpoint{1.744132in}{0.315988in}}%
\pgfpathlineto{\pgfqpoint{1.744132in}{1.563944in}}%
\pgfpathlineto{\pgfqpoint{1.599959in}{1.563944in}}%
\pgfpathlineto{\pgfqpoint{1.599959in}{0.315988in}}%
\pgfpathclose%
\pgfusepath{fill}%
\end{pgfscope}%
\begin{pgfscope}%
\pgfpathrectangle{\pgfqpoint{0.394308in}{0.315988in}}{\pgfqpoint{2.735692in}{2.055128in}}%
\pgfusepath{clip}%
\pgfsetbuttcap%
\pgfsetmiterjoin%
\definecolor{currentfill}{rgb}{0.121569,0.466667,0.705882}%
\pgfsetfillcolor{currentfill}%
\pgfsetlinewidth{0.000000pt}%
\definecolor{currentstroke}{rgb}{0.000000,0.000000,0.000000}%
\pgfsetstrokecolor{currentstroke}%
\pgfsetstrokeopacity{0.000000}%
\pgfsetdash{}{0pt}%
\pgfpathmoveto{\pgfqpoint{1.780176in}{0.315988in}}%
\pgfpathlineto{\pgfqpoint{1.924349in}{0.315988in}}%
\pgfpathlineto{\pgfqpoint{1.924349in}{1.382617in}}%
\pgfpathlineto{\pgfqpoint{1.780176in}{1.382617in}}%
\pgfpathlineto{\pgfqpoint{1.780176in}{0.315988in}}%
\pgfpathclose%
\pgfusepath{fill}%
\end{pgfscope}%
\begin{pgfscope}%
\pgfpathrectangle{\pgfqpoint{0.394308in}{0.315988in}}{\pgfqpoint{2.735692in}{2.055128in}}%
\pgfusepath{clip}%
\pgfsetbuttcap%
\pgfsetmiterjoin%
\definecolor{currentfill}{rgb}{0.121569,0.466667,0.705882}%
\pgfsetfillcolor{currentfill}%
\pgfsetlinewidth{0.000000pt}%
\definecolor{currentstroke}{rgb}{0.000000,0.000000,0.000000}%
\pgfsetstrokecolor{currentstroke}%
\pgfsetstrokeopacity{0.000000}%
\pgfsetdash{}{0pt}%
\pgfpathmoveto{\pgfqpoint{1.960393in}{0.315988in}}%
\pgfpathlineto{\pgfqpoint{2.104566in}{0.315988in}}%
\pgfpathlineto{\pgfqpoint{2.104566in}{1.227956in}}%
\pgfpathlineto{\pgfqpoint{1.960393in}{1.227956in}}%
\pgfpathlineto{\pgfqpoint{1.960393in}{0.315988in}}%
\pgfpathclose%
\pgfusepath{fill}%
\end{pgfscope}%
\begin{pgfscope}%
\pgfpathrectangle{\pgfqpoint{0.394308in}{0.315988in}}{\pgfqpoint{2.735692in}{2.055128in}}%
\pgfusepath{clip}%
\pgfsetbuttcap%
\pgfsetmiterjoin%
\definecolor{currentfill}{rgb}{0.121569,0.466667,0.705882}%
\pgfsetfillcolor{currentfill}%
\pgfsetlinewidth{0.000000pt}%
\definecolor{currentstroke}{rgb}{0.000000,0.000000,0.000000}%
\pgfsetstrokecolor{currentstroke}%
\pgfsetstrokeopacity{0.000000}%
\pgfsetdash{}{0pt}%
\pgfpathmoveto{\pgfqpoint{2.140609in}{0.315988in}}%
\pgfpathlineto{\pgfqpoint{2.284783in}{0.315988in}}%
\pgfpathlineto{\pgfqpoint{2.284783in}{1.169291in}}%
\pgfpathlineto{\pgfqpoint{2.140609in}{1.169291in}}%
\pgfpathlineto{\pgfqpoint{2.140609in}{0.315988in}}%
\pgfpathclose%
\pgfusepath{fill}%
\end{pgfscope}%
\begin{pgfscope}%
\pgfpathrectangle{\pgfqpoint{0.394308in}{0.315988in}}{\pgfqpoint{2.735692in}{2.055128in}}%
\pgfusepath{clip}%
\pgfsetbuttcap%
\pgfsetmiterjoin%
\definecolor{currentfill}{rgb}{0.121569,0.466667,0.705882}%
\pgfsetfillcolor{currentfill}%
\pgfsetlinewidth{0.000000pt}%
\definecolor{currentstroke}{rgb}{0.000000,0.000000,0.000000}%
\pgfsetstrokecolor{currentstroke}%
\pgfsetstrokeopacity{0.000000}%
\pgfsetdash{}{0pt}%
\pgfpathmoveto{\pgfqpoint{2.320826in}{0.315988in}}%
\pgfpathlineto{\pgfqpoint{2.465000in}{0.315988in}}%
\pgfpathlineto{\pgfqpoint{2.465000in}{0.934633in}}%
\pgfpathlineto{\pgfqpoint{2.320826in}{0.934633in}}%
\pgfpathlineto{\pgfqpoint{2.320826in}{0.315988in}}%
\pgfpathclose%
\pgfusepath{fill}%
\end{pgfscope}%
\begin{pgfscope}%
\pgfpathrectangle{\pgfqpoint{0.394308in}{0.315988in}}{\pgfqpoint{2.735692in}{2.055128in}}%
\pgfusepath{clip}%
\pgfsetbuttcap%
\pgfsetmiterjoin%
\definecolor{currentfill}{rgb}{0.121569,0.466667,0.705882}%
\pgfsetfillcolor{currentfill}%
\pgfsetlinewidth{0.000000pt}%
\definecolor{currentstroke}{rgb}{0.000000,0.000000,0.000000}%
\pgfsetstrokecolor{currentstroke}%
\pgfsetstrokeopacity{0.000000}%
\pgfsetdash{}{0pt}%
\pgfpathmoveto{\pgfqpoint{2.501043in}{0.315988in}}%
\pgfpathlineto{\pgfqpoint{2.645217in}{0.315988in}}%
\pgfpathlineto{\pgfqpoint{2.645217in}{0.347987in}}%
\pgfpathlineto{\pgfqpoint{2.501043in}{0.347987in}}%
\pgfpathlineto{\pgfqpoint{2.501043in}{0.315988in}}%
\pgfpathclose%
\pgfusepath{fill}%
\end{pgfscope}%
\begin{pgfscope}%
\pgfpathrectangle{\pgfqpoint{0.394308in}{0.315988in}}{\pgfqpoint{2.735692in}{2.055128in}}%
\pgfusepath{clip}%
\pgfsetbuttcap%
\pgfsetmiterjoin%
\definecolor{currentfill}{rgb}{0.121569,0.466667,0.705882}%
\pgfsetfillcolor{currentfill}%
\pgfsetlinewidth{0.000000pt}%
\definecolor{currentstroke}{rgb}{0.000000,0.000000,0.000000}%
\pgfsetstrokecolor{currentstroke}%
\pgfsetstrokeopacity{0.000000}%
\pgfsetdash{}{0pt}%
\pgfpathmoveto{\pgfqpoint{2.681260in}{0.315988in}}%
\pgfpathlineto{\pgfqpoint{2.825434in}{0.315988in}}%
\pgfpathlineto{\pgfqpoint{2.825434in}{0.347987in}}%
\pgfpathlineto{\pgfqpoint{2.681260in}{0.347987in}}%
\pgfpathlineto{\pgfqpoint{2.681260in}{0.315988in}}%
\pgfpathclose%
\pgfusepath{fill}%
\end{pgfscope}%
\begin{pgfscope}%
\pgfpathrectangle{\pgfqpoint{0.394308in}{0.315988in}}{\pgfqpoint{2.735692in}{2.055128in}}%
\pgfusepath{clip}%
\pgfsetbuttcap%
\pgfsetmiterjoin%
\definecolor{currentfill}{rgb}{0.121569,0.466667,0.705882}%
\pgfsetfillcolor{currentfill}%
\pgfsetlinewidth{0.000000pt}%
\definecolor{currentstroke}{rgb}{0.000000,0.000000,0.000000}%
\pgfsetstrokecolor{currentstroke}%
\pgfsetstrokeopacity{0.000000}%
\pgfsetdash{}{0pt}%
\pgfpathmoveto{\pgfqpoint{2.861477in}{0.315988in}}%
\pgfpathlineto{\pgfqpoint{3.005650in}{0.315988in}}%
\pgfpathlineto{\pgfqpoint{3.005650in}{0.337320in}}%
\pgfpathlineto{\pgfqpoint{2.861477in}{0.337320in}}%
\pgfpathlineto{\pgfqpoint{2.861477in}{0.315988in}}%
\pgfpathclose%
\pgfusepath{fill}%
\end{pgfscope}%
\begin{pgfscope}%
\pgfsetbuttcap%
\pgfsetroundjoin%
\definecolor{currentfill}{rgb}{0.000000,0.000000,0.000000}%
\pgfsetfillcolor{currentfill}%
\pgfsetlinewidth{0.803000pt}%
\definecolor{currentstroke}{rgb}{0.000000,0.000000,0.000000}%
\pgfsetstrokecolor{currentstroke}%
\pgfsetdash{}{0pt}%
\pgfsys@defobject{currentmarker}{\pgfqpoint{0.000000in}{-0.048611in}}{\pgfqpoint{0.000000in}{0.000000in}}{%
\pgfpathmoveto{\pgfqpoint{0.000000in}{0.000000in}}%
\pgfpathlineto{\pgfqpoint{0.000000in}{-0.048611in}}%
\pgfusepath{stroke,fill}%
}%
\begin{pgfscope}%
\pgfsys@transformshift{0.590744in}{0.315988in}%
\pgfsys@useobject{currentmarker}{}%
\end{pgfscope}%
\end{pgfscope}%
\begin{pgfscope}%
\definecolor{textcolor}{rgb}{0.000000,0.000000,0.000000}%
\pgfsetstrokecolor{textcolor}%
\pgfsetfillcolor{textcolor}%
\pgftext[x=0.590744in,y=0.218766in,,top]{\color{textcolor}\rmfamily\fontsize{8.000000}{9.600000}\selectfont A}%
\end{pgfscope}%
\begin{pgfscope}%
\pgfsetbuttcap%
\pgfsetroundjoin%
\definecolor{currentfill}{rgb}{0.000000,0.000000,0.000000}%
\pgfsetfillcolor{currentfill}%
\pgfsetlinewidth{0.803000pt}%
\definecolor{currentstroke}{rgb}{0.000000,0.000000,0.000000}%
\pgfsetstrokecolor{currentstroke}%
\pgfsetdash{}{0pt}%
\pgfsys@defobject{currentmarker}{\pgfqpoint{0.000000in}{-0.048611in}}{\pgfqpoint{0.000000in}{0.000000in}}{%
\pgfpathmoveto{\pgfqpoint{0.000000in}{0.000000in}}%
\pgfpathlineto{\pgfqpoint{0.000000in}{-0.048611in}}%
\pgfusepath{stroke,fill}%
}%
\begin{pgfscope}%
\pgfsys@transformshift{0.770961in}{0.315988in}%
\pgfsys@useobject{currentmarker}{}%
\end{pgfscope}%
\end{pgfscope}%
\begin{pgfscope}%
\definecolor{textcolor}{rgb}{0.000000,0.000000,0.000000}%
\pgfsetstrokecolor{textcolor}%
\pgfsetfillcolor{textcolor}%
\pgftext[x=0.770961in,y=0.218766in,,top]{\color{textcolor}\rmfamily\fontsize{8.000000}{9.600000}\selectfont B}%
\end{pgfscope}%
\begin{pgfscope}%
\pgfsetbuttcap%
\pgfsetroundjoin%
\definecolor{currentfill}{rgb}{0.000000,0.000000,0.000000}%
\pgfsetfillcolor{currentfill}%
\pgfsetlinewidth{0.803000pt}%
\definecolor{currentstroke}{rgb}{0.000000,0.000000,0.000000}%
\pgfsetstrokecolor{currentstroke}%
\pgfsetdash{}{0pt}%
\pgfsys@defobject{currentmarker}{\pgfqpoint{0.000000in}{-0.048611in}}{\pgfqpoint{0.000000in}{0.000000in}}{%
\pgfpathmoveto{\pgfqpoint{0.000000in}{0.000000in}}%
\pgfpathlineto{\pgfqpoint{0.000000in}{-0.048611in}}%
\pgfusepath{stroke,fill}%
}%
\begin{pgfscope}%
\pgfsys@transformshift{0.951178in}{0.315988in}%
\pgfsys@useobject{currentmarker}{}%
\end{pgfscope}%
\end{pgfscope}%
\begin{pgfscope}%
\definecolor{textcolor}{rgb}{0.000000,0.000000,0.000000}%
\pgfsetstrokecolor{textcolor}%
\pgfsetfillcolor{textcolor}%
\pgftext[x=0.951178in,y=0.218766in,,top]{\color{textcolor}\rmfamily\fontsize{8.000000}{9.600000}\selectfont C}%
\end{pgfscope}%
\begin{pgfscope}%
\pgfsetbuttcap%
\pgfsetroundjoin%
\definecolor{currentfill}{rgb}{0.000000,0.000000,0.000000}%
\pgfsetfillcolor{currentfill}%
\pgfsetlinewidth{0.803000pt}%
\definecolor{currentstroke}{rgb}{0.000000,0.000000,0.000000}%
\pgfsetstrokecolor{currentstroke}%
\pgfsetdash{}{0pt}%
\pgfsys@defobject{currentmarker}{\pgfqpoint{0.000000in}{-0.048611in}}{\pgfqpoint{0.000000in}{0.000000in}}{%
\pgfpathmoveto{\pgfqpoint{0.000000in}{0.000000in}}%
\pgfpathlineto{\pgfqpoint{0.000000in}{-0.048611in}}%
\pgfusepath{stroke,fill}%
}%
\begin{pgfscope}%
\pgfsys@transformshift{1.131395in}{0.315988in}%
\pgfsys@useobject{currentmarker}{}%
\end{pgfscope}%
\end{pgfscope}%
\begin{pgfscope}%
\definecolor{textcolor}{rgb}{0.000000,0.000000,0.000000}%
\pgfsetstrokecolor{textcolor}%
\pgfsetfillcolor{textcolor}%
\pgftext[x=1.131395in,y=0.218766in,,top]{\color{textcolor}\rmfamily\fontsize{8.000000}{9.600000}\selectfont D}%
\end{pgfscope}%
\begin{pgfscope}%
\pgfsetbuttcap%
\pgfsetroundjoin%
\definecolor{currentfill}{rgb}{0.000000,0.000000,0.000000}%
\pgfsetfillcolor{currentfill}%
\pgfsetlinewidth{0.803000pt}%
\definecolor{currentstroke}{rgb}{0.000000,0.000000,0.000000}%
\pgfsetstrokecolor{currentstroke}%
\pgfsetdash{}{0pt}%
\pgfsys@defobject{currentmarker}{\pgfqpoint{0.000000in}{-0.048611in}}{\pgfqpoint{0.000000in}{0.000000in}}{%
\pgfpathmoveto{\pgfqpoint{0.000000in}{0.000000in}}%
\pgfpathlineto{\pgfqpoint{0.000000in}{-0.048611in}}%
\pgfusepath{stroke,fill}%
}%
\begin{pgfscope}%
\pgfsys@transformshift{1.311612in}{0.315988in}%
\pgfsys@useobject{currentmarker}{}%
\end{pgfscope}%
\end{pgfscope}%
\begin{pgfscope}%
\definecolor{textcolor}{rgb}{0.000000,0.000000,0.000000}%
\pgfsetstrokecolor{textcolor}%
\pgfsetfillcolor{textcolor}%
\pgftext[x=1.311612in,y=0.218766in,,top]{\color{textcolor}\rmfamily\fontsize{8.000000}{9.600000}\selectfont E}%
\end{pgfscope}%
\begin{pgfscope}%
\pgfsetbuttcap%
\pgfsetroundjoin%
\definecolor{currentfill}{rgb}{0.000000,0.000000,0.000000}%
\pgfsetfillcolor{currentfill}%
\pgfsetlinewidth{0.803000pt}%
\definecolor{currentstroke}{rgb}{0.000000,0.000000,0.000000}%
\pgfsetstrokecolor{currentstroke}%
\pgfsetdash{}{0pt}%
\pgfsys@defobject{currentmarker}{\pgfqpoint{0.000000in}{-0.048611in}}{\pgfqpoint{0.000000in}{0.000000in}}{%
\pgfpathmoveto{\pgfqpoint{0.000000in}{0.000000in}}%
\pgfpathlineto{\pgfqpoint{0.000000in}{-0.048611in}}%
\pgfusepath{stroke,fill}%
}%
\begin{pgfscope}%
\pgfsys@transformshift{1.491829in}{0.315988in}%
\pgfsys@useobject{currentmarker}{}%
\end{pgfscope}%
\end{pgfscope}%
\begin{pgfscope}%
\definecolor{textcolor}{rgb}{0.000000,0.000000,0.000000}%
\pgfsetstrokecolor{textcolor}%
\pgfsetfillcolor{textcolor}%
\pgftext[x=1.491829in,y=0.218766in,,top]{\color{textcolor}\rmfamily\fontsize{8.000000}{9.600000}\selectfont F}%
\end{pgfscope}%
\begin{pgfscope}%
\pgfsetbuttcap%
\pgfsetroundjoin%
\definecolor{currentfill}{rgb}{0.000000,0.000000,0.000000}%
\pgfsetfillcolor{currentfill}%
\pgfsetlinewidth{0.803000pt}%
\definecolor{currentstroke}{rgb}{0.000000,0.000000,0.000000}%
\pgfsetstrokecolor{currentstroke}%
\pgfsetdash{}{0pt}%
\pgfsys@defobject{currentmarker}{\pgfqpoint{0.000000in}{-0.048611in}}{\pgfqpoint{0.000000in}{0.000000in}}{%
\pgfpathmoveto{\pgfqpoint{0.000000in}{0.000000in}}%
\pgfpathlineto{\pgfqpoint{0.000000in}{-0.048611in}}%
\pgfusepath{stroke,fill}%
}%
\begin{pgfscope}%
\pgfsys@transformshift{1.672046in}{0.315988in}%
\pgfsys@useobject{currentmarker}{}%
\end{pgfscope}%
\end{pgfscope}%
\begin{pgfscope}%
\definecolor{textcolor}{rgb}{0.000000,0.000000,0.000000}%
\pgfsetstrokecolor{textcolor}%
\pgfsetfillcolor{textcolor}%
\pgftext[x=1.672046in,y=0.218766in,,top]{\color{textcolor}\rmfamily\fontsize{8.000000}{9.600000}\selectfont G}%
\end{pgfscope}%
\begin{pgfscope}%
\pgfsetbuttcap%
\pgfsetroundjoin%
\definecolor{currentfill}{rgb}{0.000000,0.000000,0.000000}%
\pgfsetfillcolor{currentfill}%
\pgfsetlinewidth{0.803000pt}%
\definecolor{currentstroke}{rgb}{0.000000,0.000000,0.000000}%
\pgfsetstrokecolor{currentstroke}%
\pgfsetdash{}{0pt}%
\pgfsys@defobject{currentmarker}{\pgfqpoint{0.000000in}{-0.048611in}}{\pgfqpoint{0.000000in}{0.000000in}}{%
\pgfpathmoveto{\pgfqpoint{0.000000in}{0.000000in}}%
\pgfpathlineto{\pgfqpoint{0.000000in}{-0.048611in}}%
\pgfusepath{stroke,fill}%
}%
\begin{pgfscope}%
\pgfsys@transformshift{1.852262in}{0.315988in}%
\pgfsys@useobject{currentmarker}{}%
\end{pgfscope}%
\end{pgfscope}%
\begin{pgfscope}%
\definecolor{textcolor}{rgb}{0.000000,0.000000,0.000000}%
\pgfsetstrokecolor{textcolor}%
\pgfsetfillcolor{textcolor}%
\pgftext[x=1.852262in,y=0.218766in,,top]{\color{textcolor}\rmfamily\fontsize{8.000000}{9.600000}\selectfont H}%
\end{pgfscope}%
\begin{pgfscope}%
\pgfsetbuttcap%
\pgfsetroundjoin%
\definecolor{currentfill}{rgb}{0.000000,0.000000,0.000000}%
\pgfsetfillcolor{currentfill}%
\pgfsetlinewidth{0.803000pt}%
\definecolor{currentstroke}{rgb}{0.000000,0.000000,0.000000}%
\pgfsetstrokecolor{currentstroke}%
\pgfsetdash{}{0pt}%
\pgfsys@defobject{currentmarker}{\pgfqpoint{0.000000in}{-0.048611in}}{\pgfqpoint{0.000000in}{0.000000in}}{%
\pgfpathmoveto{\pgfqpoint{0.000000in}{0.000000in}}%
\pgfpathlineto{\pgfqpoint{0.000000in}{-0.048611in}}%
\pgfusepath{stroke,fill}%
}%
\begin{pgfscope}%
\pgfsys@transformshift{2.032479in}{0.315988in}%
\pgfsys@useobject{currentmarker}{}%
\end{pgfscope}%
\end{pgfscope}%
\begin{pgfscope}%
\definecolor{textcolor}{rgb}{0.000000,0.000000,0.000000}%
\pgfsetstrokecolor{textcolor}%
\pgfsetfillcolor{textcolor}%
\pgftext[x=2.032479in,y=0.218766in,,top]{\color{textcolor}\rmfamily\fontsize{8.000000}{9.600000}\selectfont I}%
\end{pgfscope}%
\begin{pgfscope}%
\pgfsetbuttcap%
\pgfsetroundjoin%
\definecolor{currentfill}{rgb}{0.000000,0.000000,0.000000}%
\pgfsetfillcolor{currentfill}%
\pgfsetlinewidth{0.803000pt}%
\definecolor{currentstroke}{rgb}{0.000000,0.000000,0.000000}%
\pgfsetstrokecolor{currentstroke}%
\pgfsetdash{}{0pt}%
\pgfsys@defobject{currentmarker}{\pgfqpoint{0.000000in}{-0.048611in}}{\pgfqpoint{0.000000in}{0.000000in}}{%
\pgfpathmoveto{\pgfqpoint{0.000000in}{0.000000in}}%
\pgfpathlineto{\pgfqpoint{0.000000in}{-0.048611in}}%
\pgfusepath{stroke,fill}%
}%
\begin{pgfscope}%
\pgfsys@transformshift{2.212696in}{0.315988in}%
\pgfsys@useobject{currentmarker}{}%
\end{pgfscope}%
\end{pgfscope}%
\begin{pgfscope}%
\definecolor{textcolor}{rgb}{0.000000,0.000000,0.000000}%
\pgfsetstrokecolor{textcolor}%
\pgfsetfillcolor{textcolor}%
\pgftext[x=2.212696in,y=0.218766in,,top]{\color{textcolor}\rmfamily\fontsize{8.000000}{9.600000}\selectfont J}%
\end{pgfscope}%
\begin{pgfscope}%
\pgfsetbuttcap%
\pgfsetroundjoin%
\definecolor{currentfill}{rgb}{0.000000,0.000000,0.000000}%
\pgfsetfillcolor{currentfill}%
\pgfsetlinewidth{0.803000pt}%
\definecolor{currentstroke}{rgb}{0.000000,0.000000,0.000000}%
\pgfsetstrokecolor{currentstroke}%
\pgfsetdash{}{0pt}%
\pgfsys@defobject{currentmarker}{\pgfqpoint{0.000000in}{-0.048611in}}{\pgfqpoint{0.000000in}{0.000000in}}{%
\pgfpathmoveto{\pgfqpoint{0.000000in}{0.000000in}}%
\pgfpathlineto{\pgfqpoint{0.000000in}{-0.048611in}}%
\pgfusepath{stroke,fill}%
}%
\begin{pgfscope}%
\pgfsys@transformshift{2.392913in}{0.315988in}%
\pgfsys@useobject{currentmarker}{}%
\end{pgfscope}%
\end{pgfscope}%
\begin{pgfscope}%
\definecolor{textcolor}{rgb}{0.000000,0.000000,0.000000}%
\pgfsetstrokecolor{textcolor}%
\pgfsetfillcolor{textcolor}%
\pgftext[x=2.392913in,y=0.218766in,,top]{\color{textcolor}\rmfamily\fontsize{8.000000}{9.600000}\selectfont K}%
\end{pgfscope}%
\begin{pgfscope}%
\pgfsetbuttcap%
\pgfsetroundjoin%
\definecolor{currentfill}{rgb}{0.000000,0.000000,0.000000}%
\pgfsetfillcolor{currentfill}%
\pgfsetlinewidth{0.803000pt}%
\definecolor{currentstroke}{rgb}{0.000000,0.000000,0.000000}%
\pgfsetstrokecolor{currentstroke}%
\pgfsetdash{}{0pt}%
\pgfsys@defobject{currentmarker}{\pgfqpoint{0.000000in}{-0.048611in}}{\pgfqpoint{0.000000in}{0.000000in}}{%
\pgfpathmoveto{\pgfqpoint{0.000000in}{0.000000in}}%
\pgfpathlineto{\pgfqpoint{0.000000in}{-0.048611in}}%
\pgfusepath{stroke,fill}%
}%
\begin{pgfscope}%
\pgfsys@transformshift{2.573130in}{0.315988in}%
\pgfsys@useobject{currentmarker}{}%
\end{pgfscope}%
\end{pgfscope}%
\begin{pgfscope}%
\definecolor{textcolor}{rgb}{0.000000,0.000000,0.000000}%
\pgfsetstrokecolor{textcolor}%
\pgfsetfillcolor{textcolor}%
\pgftext[x=2.573130in,y=0.218766in,,top]{\color{textcolor}\rmfamily\fontsize{8.000000}{9.600000}\selectfont L}%
\end{pgfscope}%
\begin{pgfscope}%
\pgfsetbuttcap%
\pgfsetroundjoin%
\definecolor{currentfill}{rgb}{0.000000,0.000000,0.000000}%
\pgfsetfillcolor{currentfill}%
\pgfsetlinewidth{0.803000pt}%
\definecolor{currentstroke}{rgb}{0.000000,0.000000,0.000000}%
\pgfsetstrokecolor{currentstroke}%
\pgfsetdash{}{0pt}%
\pgfsys@defobject{currentmarker}{\pgfqpoint{0.000000in}{-0.048611in}}{\pgfqpoint{0.000000in}{0.000000in}}{%
\pgfpathmoveto{\pgfqpoint{0.000000in}{0.000000in}}%
\pgfpathlineto{\pgfqpoint{0.000000in}{-0.048611in}}%
\pgfusepath{stroke,fill}%
}%
\begin{pgfscope}%
\pgfsys@transformshift{2.753347in}{0.315988in}%
\pgfsys@useobject{currentmarker}{}%
\end{pgfscope}%
\end{pgfscope}%
\begin{pgfscope}%
\definecolor{textcolor}{rgb}{0.000000,0.000000,0.000000}%
\pgfsetstrokecolor{textcolor}%
\pgfsetfillcolor{textcolor}%
\pgftext[x=2.753347in,y=0.218766in,,top]{\color{textcolor}\rmfamily\fontsize{8.000000}{9.600000}\selectfont M}%
\end{pgfscope}%
\begin{pgfscope}%
\pgfsetbuttcap%
\pgfsetroundjoin%
\definecolor{currentfill}{rgb}{0.000000,0.000000,0.000000}%
\pgfsetfillcolor{currentfill}%
\pgfsetlinewidth{0.803000pt}%
\definecolor{currentstroke}{rgb}{0.000000,0.000000,0.000000}%
\pgfsetstrokecolor{currentstroke}%
\pgfsetdash{}{0pt}%
\pgfsys@defobject{currentmarker}{\pgfqpoint{0.000000in}{-0.048611in}}{\pgfqpoint{0.000000in}{0.000000in}}{%
\pgfpathmoveto{\pgfqpoint{0.000000in}{0.000000in}}%
\pgfpathlineto{\pgfqpoint{0.000000in}{-0.048611in}}%
\pgfusepath{stroke,fill}%
}%
\begin{pgfscope}%
\pgfsys@transformshift{2.933564in}{0.315988in}%
\pgfsys@useobject{currentmarker}{}%
\end{pgfscope}%
\end{pgfscope}%
\begin{pgfscope}%
\definecolor{textcolor}{rgb}{0.000000,0.000000,0.000000}%
\pgfsetstrokecolor{textcolor}%
\pgfsetfillcolor{textcolor}%
\pgftext[x=2.933564in,y=0.218766in,,top]{\color{textcolor}\rmfamily\fontsize{8.000000}{9.600000}\selectfont N}%
\end{pgfscope}%
\begin{pgfscope}%
\pgfsetbuttcap%
\pgfsetroundjoin%
\definecolor{currentfill}{rgb}{0.000000,0.000000,0.000000}%
\pgfsetfillcolor{currentfill}%
\pgfsetlinewidth{0.803000pt}%
\definecolor{currentstroke}{rgb}{0.000000,0.000000,0.000000}%
\pgfsetstrokecolor{currentstroke}%
\pgfsetdash{}{0pt}%
\pgfsys@defobject{currentmarker}{\pgfqpoint{-0.048611in}{0.000000in}}{\pgfqpoint{-0.000000in}{0.000000in}}{%
\pgfpathmoveto{\pgfqpoint{-0.000000in}{0.000000in}}%
\pgfpathlineto{\pgfqpoint{-0.048611in}{0.000000in}}%
\pgfusepath{stroke,fill}%
}%
\begin{pgfscope}%
\pgfsys@transformshift{0.394308in}{0.315988in}%
\pgfsys@useobject{currentmarker}{}%
\end{pgfscope}%
\end{pgfscope}%
\begin{pgfscope}%
\definecolor{textcolor}{rgb}{0.000000,0.000000,0.000000}%
\pgfsetstrokecolor{textcolor}%
\pgfsetfillcolor{textcolor}%
\pgftext[x=0.238057in, y=0.277408in, left, base]{\color{textcolor}\rmfamily\fontsize{8.000000}{9.600000}\selectfont \(\displaystyle {0}\)}%
\end{pgfscope}%
\begin{pgfscope}%
\pgfsetbuttcap%
\pgfsetroundjoin%
\definecolor{currentfill}{rgb}{0.000000,0.000000,0.000000}%
\pgfsetfillcolor{currentfill}%
\pgfsetlinewidth{0.803000pt}%
\definecolor{currentstroke}{rgb}{0.000000,0.000000,0.000000}%
\pgfsetstrokecolor{currentstroke}%
\pgfsetdash{}{0pt}%
\pgfsys@defobject{currentmarker}{\pgfqpoint{-0.048611in}{0.000000in}}{\pgfqpoint{-0.000000in}{0.000000in}}{%
\pgfpathmoveto{\pgfqpoint{-0.000000in}{0.000000in}}%
\pgfpathlineto{\pgfqpoint{-0.048611in}{0.000000in}}%
\pgfusepath{stroke,fill}%
}%
\begin{pgfscope}%
\pgfsys@transformshift{0.394308in}{0.582645in}%
\pgfsys@useobject{currentmarker}{}%
\end{pgfscope}%
\end{pgfscope}%
\begin{pgfscope}%
\definecolor{textcolor}{rgb}{0.000000,0.000000,0.000000}%
\pgfsetstrokecolor{textcolor}%
\pgfsetfillcolor{textcolor}%
\pgftext[x=0.120000in, y=0.544065in, left, base]{\color{textcolor}\rmfamily\fontsize{8.000000}{9.600000}\selectfont \(\displaystyle {100}\)}%
\end{pgfscope}%
\begin{pgfscope}%
\pgfsetbuttcap%
\pgfsetroundjoin%
\definecolor{currentfill}{rgb}{0.000000,0.000000,0.000000}%
\pgfsetfillcolor{currentfill}%
\pgfsetlinewidth{0.803000pt}%
\definecolor{currentstroke}{rgb}{0.000000,0.000000,0.000000}%
\pgfsetstrokecolor{currentstroke}%
\pgfsetdash{}{0pt}%
\pgfsys@defobject{currentmarker}{\pgfqpoint{-0.048611in}{0.000000in}}{\pgfqpoint{-0.000000in}{0.000000in}}{%
\pgfpathmoveto{\pgfqpoint{-0.000000in}{0.000000in}}%
\pgfpathlineto{\pgfqpoint{-0.048611in}{0.000000in}}%
\pgfusepath{stroke,fill}%
}%
\begin{pgfscope}%
\pgfsys@transformshift{0.394308in}{0.849302in}%
\pgfsys@useobject{currentmarker}{}%
\end{pgfscope}%
\end{pgfscope}%
\begin{pgfscope}%
\definecolor{textcolor}{rgb}{0.000000,0.000000,0.000000}%
\pgfsetstrokecolor{textcolor}%
\pgfsetfillcolor{textcolor}%
\pgftext[x=0.120000in, y=0.810722in, left, base]{\color{textcolor}\rmfamily\fontsize{8.000000}{9.600000}\selectfont \(\displaystyle {200}\)}%
\end{pgfscope}%
\begin{pgfscope}%
\pgfsetbuttcap%
\pgfsetroundjoin%
\definecolor{currentfill}{rgb}{0.000000,0.000000,0.000000}%
\pgfsetfillcolor{currentfill}%
\pgfsetlinewidth{0.803000pt}%
\definecolor{currentstroke}{rgb}{0.000000,0.000000,0.000000}%
\pgfsetstrokecolor{currentstroke}%
\pgfsetdash{}{0pt}%
\pgfsys@defobject{currentmarker}{\pgfqpoint{-0.048611in}{0.000000in}}{\pgfqpoint{-0.000000in}{0.000000in}}{%
\pgfpathmoveto{\pgfqpoint{-0.000000in}{0.000000in}}%
\pgfpathlineto{\pgfqpoint{-0.048611in}{0.000000in}}%
\pgfusepath{stroke,fill}%
}%
\begin{pgfscope}%
\pgfsys@transformshift{0.394308in}{1.115960in}%
\pgfsys@useobject{currentmarker}{}%
\end{pgfscope}%
\end{pgfscope}%
\begin{pgfscope}%
\definecolor{textcolor}{rgb}{0.000000,0.000000,0.000000}%
\pgfsetstrokecolor{textcolor}%
\pgfsetfillcolor{textcolor}%
\pgftext[x=0.120000in, y=1.077379in, left, base]{\color{textcolor}\rmfamily\fontsize{8.000000}{9.600000}\selectfont \(\displaystyle {300}\)}%
\end{pgfscope}%
\begin{pgfscope}%
\pgfsetbuttcap%
\pgfsetroundjoin%
\definecolor{currentfill}{rgb}{0.000000,0.000000,0.000000}%
\pgfsetfillcolor{currentfill}%
\pgfsetlinewidth{0.803000pt}%
\definecolor{currentstroke}{rgb}{0.000000,0.000000,0.000000}%
\pgfsetstrokecolor{currentstroke}%
\pgfsetdash{}{0pt}%
\pgfsys@defobject{currentmarker}{\pgfqpoint{-0.048611in}{0.000000in}}{\pgfqpoint{-0.000000in}{0.000000in}}{%
\pgfpathmoveto{\pgfqpoint{-0.000000in}{0.000000in}}%
\pgfpathlineto{\pgfqpoint{-0.048611in}{0.000000in}}%
\pgfusepath{stroke,fill}%
}%
\begin{pgfscope}%
\pgfsys@transformshift{0.394308in}{1.382617in}%
\pgfsys@useobject{currentmarker}{}%
\end{pgfscope}%
\end{pgfscope}%
\begin{pgfscope}%
\definecolor{textcolor}{rgb}{0.000000,0.000000,0.000000}%
\pgfsetstrokecolor{textcolor}%
\pgfsetfillcolor{textcolor}%
\pgftext[x=0.120000in, y=1.344037in, left, base]{\color{textcolor}\rmfamily\fontsize{8.000000}{9.600000}\selectfont \(\displaystyle {400}\)}%
\end{pgfscope}%
\begin{pgfscope}%
\pgfsetbuttcap%
\pgfsetroundjoin%
\definecolor{currentfill}{rgb}{0.000000,0.000000,0.000000}%
\pgfsetfillcolor{currentfill}%
\pgfsetlinewidth{0.803000pt}%
\definecolor{currentstroke}{rgb}{0.000000,0.000000,0.000000}%
\pgfsetstrokecolor{currentstroke}%
\pgfsetdash{}{0pt}%
\pgfsys@defobject{currentmarker}{\pgfqpoint{-0.048611in}{0.000000in}}{\pgfqpoint{-0.000000in}{0.000000in}}{%
\pgfpathmoveto{\pgfqpoint{-0.000000in}{0.000000in}}%
\pgfpathlineto{\pgfqpoint{-0.048611in}{0.000000in}}%
\pgfusepath{stroke,fill}%
}%
\begin{pgfscope}%
\pgfsys@transformshift{0.394308in}{1.649274in}%
\pgfsys@useobject{currentmarker}{}%
\end{pgfscope}%
\end{pgfscope}%
\begin{pgfscope}%
\definecolor{textcolor}{rgb}{0.000000,0.000000,0.000000}%
\pgfsetstrokecolor{textcolor}%
\pgfsetfillcolor{textcolor}%
\pgftext[x=0.120000in, y=1.610694in, left, base]{\color{textcolor}\rmfamily\fontsize{8.000000}{9.600000}\selectfont \(\displaystyle {500}\)}%
\end{pgfscope}%
\begin{pgfscope}%
\pgfsetbuttcap%
\pgfsetroundjoin%
\definecolor{currentfill}{rgb}{0.000000,0.000000,0.000000}%
\pgfsetfillcolor{currentfill}%
\pgfsetlinewidth{0.803000pt}%
\definecolor{currentstroke}{rgb}{0.000000,0.000000,0.000000}%
\pgfsetstrokecolor{currentstroke}%
\pgfsetdash{}{0pt}%
\pgfsys@defobject{currentmarker}{\pgfqpoint{-0.048611in}{0.000000in}}{\pgfqpoint{-0.000000in}{0.000000in}}{%
\pgfpathmoveto{\pgfqpoint{-0.000000in}{0.000000in}}%
\pgfpathlineto{\pgfqpoint{-0.048611in}{0.000000in}}%
\pgfusepath{stroke,fill}%
}%
\begin{pgfscope}%
\pgfsys@transformshift{0.394308in}{1.915931in}%
\pgfsys@useobject{currentmarker}{}%
\end{pgfscope}%
\end{pgfscope}%
\begin{pgfscope}%
\definecolor{textcolor}{rgb}{0.000000,0.000000,0.000000}%
\pgfsetstrokecolor{textcolor}%
\pgfsetfillcolor{textcolor}%
\pgftext[x=0.120000in, y=1.877351in, left, base]{\color{textcolor}\rmfamily\fontsize{8.000000}{9.600000}\selectfont \(\displaystyle {600}\)}%
\end{pgfscope}%
\begin{pgfscope}%
\pgfsetbuttcap%
\pgfsetroundjoin%
\definecolor{currentfill}{rgb}{0.000000,0.000000,0.000000}%
\pgfsetfillcolor{currentfill}%
\pgfsetlinewidth{0.803000pt}%
\definecolor{currentstroke}{rgb}{0.000000,0.000000,0.000000}%
\pgfsetstrokecolor{currentstroke}%
\pgfsetdash{}{0pt}%
\pgfsys@defobject{currentmarker}{\pgfqpoint{-0.048611in}{0.000000in}}{\pgfqpoint{-0.000000in}{0.000000in}}{%
\pgfpathmoveto{\pgfqpoint{-0.000000in}{0.000000in}}%
\pgfpathlineto{\pgfqpoint{-0.048611in}{0.000000in}}%
\pgfusepath{stroke,fill}%
}%
\begin{pgfscope}%
\pgfsys@transformshift{0.394308in}{2.182589in}%
\pgfsys@useobject{currentmarker}{}%
\end{pgfscope}%
\end{pgfscope}%
\begin{pgfscope}%
\definecolor{textcolor}{rgb}{0.000000,0.000000,0.000000}%
\pgfsetstrokecolor{textcolor}%
\pgfsetfillcolor{textcolor}%
\pgftext[x=0.120000in, y=2.144008in, left, base]{\color{textcolor}\rmfamily\fontsize{8.000000}{9.600000}\selectfont \(\displaystyle {700}\)}%
\end{pgfscope}%
\begin{pgfscope}%
\pgfsetrectcap%
\pgfsetmiterjoin%
\pgfsetlinewidth{0.803000pt}%
\definecolor{currentstroke}{rgb}{0.000000,0.000000,0.000000}%
\pgfsetstrokecolor{currentstroke}%
\pgfsetdash{}{0pt}%
\pgfpathmoveto{\pgfqpoint{0.394308in}{0.315988in}}%
\pgfpathlineto{\pgfqpoint{0.394308in}{2.371115in}}%
\pgfusepath{stroke}%
\end{pgfscope}%
\begin{pgfscope}%
\pgfsetrectcap%
\pgfsetmiterjoin%
\pgfsetlinewidth{0.803000pt}%
\definecolor{currentstroke}{rgb}{0.000000,0.000000,0.000000}%
\pgfsetstrokecolor{currentstroke}%
\pgfsetdash{}{0pt}%
\pgfpathmoveto{\pgfqpoint{3.130000in}{0.315988in}}%
\pgfpathlineto{\pgfqpoint{3.130000in}{2.371115in}}%
\pgfusepath{stroke}%
\end{pgfscope}%
\begin{pgfscope}%
\pgfsetrectcap%
\pgfsetmiterjoin%
\pgfsetlinewidth{0.803000pt}%
\definecolor{currentstroke}{rgb}{0.000000,0.000000,0.000000}%
\pgfsetstrokecolor{currentstroke}%
\pgfsetdash{}{0pt}%
\pgfpathmoveto{\pgfqpoint{0.394308in}{0.315988in}}%
\pgfpathlineto{\pgfqpoint{3.130000in}{0.315988in}}%
\pgfusepath{stroke}%
\end{pgfscope}%
\begin{pgfscope}%
\pgfsetrectcap%
\pgfsetmiterjoin%
\pgfsetlinewidth{0.803000pt}%
\definecolor{currentstroke}{rgb}{0.000000,0.000000,0.000000}%
\pgfsetstrokecolor{currentstroke}%
\pgfsetdash{}{0pt}%
\pgfpathmoveto{\pgfqpoint{0.394308in}{2.371115in}}%
\pgfpathlineto{\pgfqpoint{3.130000in}{2.371115in}}%
\pgfusepath{stroke}%
\end{pgfscope}%
\begin{pgfscope}%
\definecolor{textcolor}{rgb}{0.000000,0.000000,0.000000}%
\pgfsetstrokecolor{textcolor}%
\pgfsetfillcolor{textcolor}%
\pgftext[x=0.590744in,y=2.273252in,,bottom]{\color{textcolor}\rmfamily\fontsize{8.000000}{9.600000}\selectfont 734}%
\end{pgfscope}%
\begin{pgfscope}%
\definecolor{textcolor}{rgb}{0.000000,0.000000,0.000000}%
\pgfsetstrokecolor{textcolor}%
\pgfsetfillcolor{textcolor}%
\pgftext[x=0.770961in,y=1.995929in,,bottom]{\color{textcolor}\rmfamily\fontsize{8.000000}{9.600000}\selectfont 630}%
\end{pgfscope}%
\begin{pgfscope}%
\definecolor{textcolor}{rgb}{0.000000,0.000000,0.000000}%
\pgfsetstrokecolor{textcolor}%
\pgfsetfillcolor{textcolor}%
\pgftext[x=0.951178in,y=1.979929in,,bottom]{\color{textcolor}\rmfamily\fontsize{8.000000}{9.600000}\selectfont 624}%
\end{pgfscope}%
\begin{pgfscope}%
\definecolor{textcolor}{rgb}{0.000000,0.000000,0.000000}%
\pgfsetstrokecolor{textcolor}%
\pgfsetfillcolor{textcolor}%
\pgftext[x=1.131395in,y=1.953263in,,bottom]{\color{textcolor}\rmfamily\fontsize{8.000000}{9.600000}\selectfont 614}%
\end{pgfscope}%
\begin{pgfscope}%
\definecolor{textcolor}{rgb}{0.000000,0.000000,0.000000}%
\pgfsetstrokecolor{textcolor}%
\pgfsetfillcolor{textcolor}%
\pgftext[x=1.311612in,y=1.825268in,,bottom]{\color{textcolor}\rmfamily\fontsize{8.000000}{9.600000}\selectfont 566}%
\end{pgfscope}%
\begin{pgfscope}%
\definecolor{textcolor}{rgb}{0.000000,0.000000,0.000000}%
\pgfsetstrokecolor{textcolor}%
\pgfsetfillcolor{textcolor}%
\pgftext[x=1.491829in,y=1.579943in,,bottom]{\color{textcolor}\rmfamily\fontsize{8.000000}{9.600000}\selectfont 474}%
\end{pgfscope}%
\begin{pgfscope}%
\definecolor{textcolor}{rgb}{0.000000,0.000000,0.000000}%
\pgfsetstrokecolor{textcolor}%
\pgfsetfillcolor{textcolor}%
\pgftext[x=1.672046in,y=1.563944in,,bottom]{\color{textcolor}\rmfamily\fontsize{8.000000}{9.600000}\selectfont 468}%
\end{pgfscope}%
\begin{pgfscope}%
\definecolor{textcolor}{rgb}{0.000000,0.000000,0.000000}%
\pgfsetstrokecolor{textcolor}%
\pgfsetfillcolor{textcolor}%
\pgftext[x=1.852262in,y=1.382617in,,bottom]{\color{textcolor}\rmfamily\fontsize{8.000000}{9.600000}\selectfont 400}%
\end{pgfscope}%
\begin{pgfscope}%
\definecolor{textcolor}{rgb}{0.000000,0.000000,0.000000}%
\pgfsetstrokecolor{textcolor}%
\pgfsetfillcolor{textcolor}%
\pgftext[x=2.032479in,y=1.227956in,,bottom]{\color{textcolor}\rmfamily\fontsize{8.000000}{9.600000}\selectfont 342}%
\end{pgfscope}%
\begin{pgfscope}%
\definecolor{textcolor}{rgb}{0.000000,0.000000,0.000000}%
\pgfsetstrokecolor{textcolor}%
\pgfsetfillcolor{textcolor}%
\pgftext[x=2.212696in,y=1.169291in,,bottom]{\color{textcolor}\rmfamily\fontsize{8.000000}{9.600000}\selectfont 320}%
\end{pgfscope}%
\begin{pgfscope}%
\definecolor{textcolor}{rgb}{0.000000,0.000000,0.000000}%
\pgfsetstrokecolor{textcolor}%
\pgfsetfillcolor{textcolor}%
\pgftext[x=2.392913in,y=0.934633in,,bottom]{\color{textcolor}\rmfamily\fontsize{8.000000}{9.600000}\selectfont 232}%
\end{pgfscope}%
\begin{pgfscope}%
\definecolor{textcolor}{rgb}{0.000000,0.000000,0.000000}%
\pgfsetstrokecolor{textcolor}%
\pgfsetfillcolor{textcolor}%
\pgftext[x=2.573130in,y=0.347987in,,bottom]{\color{textcolor}\rmfamily\fontsize{8.000000}{9.600000}\selectfont 12}%
\end{pgfscope}%
\begin{pgfscope}%
\definecolor{textcolor}{rgb}{0.000000,0.000000,0.000000}%
\pgfsetstrokecolor{textcolor}%
\pgfsetfillcolor{textcolor}%
\pgftext[x=2.753347in,y=0.347987in,,bottom]{\color{textcolor}\rmfamily\fontsize{8.000000}{9.600000}\selectfont 12}%
\end{pgfscope}%
\begin{pgfscope}%
\definecolor{textcolor}{rgb}{0.000000,0.000000,0.000000}%
\pgfsetstrokecolor{textcolor}%
\pgfsetfillcolor{textcolor}%
\pgftext[x=2.933564in,y=0.337320in,,bottom]{\color{textcolor}\rmfamily\fontsize{8.000000}{9.600000}\selectfont 8}%
\end{pgfscope}%
\end{pgfpicture}%
\makeatother%
\endgroup%